\pdfoutput=1
\documentclass{article} %
\usepackage{iclr2027_conference,times}

\usepackage{amsmath,amsfonts,bm}

\def\Figref#1{Figure~\ref{#1}}

\def\secref#1{section~\ref{#1}}

\def\eqref#1{equation~\ref{#1}}

\def\1{\bm{1}}

\DeclareMathAlphabet{\mathsfit}{\encodingdefault}{\sfdefault}{m}{sl}
\SetMathAlphabet{\mathsfit}{bold}{\encodingdefault}{\sfdefault}{bx}{n}

\usepackage{hyperref}
\usepackage{url}

\usepackage{algorithm}
\usepackage{algorithmic}

\usepackage{amsmath}
\usepackage{amssymb}
\usepackage{mathtools}
\usepackage{amsthm}

\usepackage{multirow}
\usepackage{tabularx}
\usepackage{makecell}
\usepackage{caption}

\usepackage[table]{xcolor}
\definecolor{Gray1}{gray}{0.9}
\definecolor{Gain}{RGB}{0,90,200}

\newcommand{\revb}[1]{#1} %

\usepackage[table]{xcolor}

\providecommand{\secref}[1]{Sec.~\ref{#1}}
\newcommand{\eqnref}[1]{Eq.~(\ref{#1})}

\newcommand{\Tabref}[1]{Table~\ref{#1}}

\newcommand{\ie}{\textit{i.e.}}
\newcommand{\eg}{\textit{e.g.}}

\DeclareMathOperator{\flatten}{flatten}
\DeclareMathOperator{\proj}{proj}

\title{A Unified Framework for EEG–Video Emotion Recognition with Brain Anatomy and Temporal Priors}

\author{
JangHyun Kim$^{1*}$, Seongro Yoon$^{2*\dagger}$, Mingyun Jeong$^{3}$, Temo Saghinadze$^{4}$, \\
\textbf{Aowen Shi$^{2}$, Donghyeon Cho$^{3}$, Fran\c{c}ois Br\'emond$^{2}$, Jinsun Park$^{1\dagger}$} \\
$^{1}$Pusan National University \quad $^{2}$Inria \quad $^{3}$Hanyang University \quad $^{4}$Georgian Technical University \\
$^{*}$Primary authors \quad $^{\dagger}$Corresponding authors \\
\texttt{jangjoa41@pusan.ac.kr, seong-ro.yoon@inria.fr}
}

\iclrfinalcopy
\begin{document}

\maketitle
\lhead{}
\renewcommand{\headrulewidth}{0pt}

\begin{abstract}
Recent studies in video- and EEG-based emotion recognition have shown notable progress. However, multi-modal emotion recognition remains largely unexplored, particularly the integration of physiological signals with video.
This integration is crucial, as EEG–video fusion combines observable behavioral cues with internal neural dynamics and enables a more comprehensive and robust characterization of human emotion.
To this end, we propose EVER, a novel EEG–Video Emotion Recognition framework that \revb{integrates information} from both modalities.
Specifically, EVER models the anatomical organization of multi-channel EEG signals and the temporal relationships among video clips through graph-based representations. These structured features are subsequently integrated with global modality representation using an inter-modal graph for emotion prediction.
To provide a comprehensive evaluation, we establish a benchmark across two public EEG-video paired datasets, Emognition and MDMER. We evaluate 12 representative models, consisting of 5 EEG-only, 5 video-only, and 2 EEG-video models.
\revb{Experiments show that EVER achieves the highest average accuracy among the evaluated models on both datasets by jointly modeling behavioral cues from video and physiological responses.}
\end{abstract}

\section{Introduction}
\label{sec:Introduction}
Emotion recognition is a fundamental task in numerous real-world applications, including mental health monitoring~\citep{aina2024hybrid}, human-computer interaction~\citep{chowdary2023deep}, and affective computing~\citep{cortinas2023toward}.
Recent advances leveraging video~\citep{zhang2021real, ben2021video} or EEG signals~\citep{peng2022joint, liu2024vbh} have demonstrated the potential of each modality to provide valuable insights into human emotion. Video captures observable behavioral cues such as facial expressions~\revb{\citep{li2020deep}}, whereas EEG reflects neural dynamics associated with internal states~\citep{lee2024eav}.
Despite their complementary characteristics, these modalities have typically been studied in isolation, and most existing datasets and benchmarks~\citep{koelstra2011deap, zheng2015investigating, zafeiriou2017aff, katsigiannis2017dreamer} are designed for unimodal learning.

However, relying on a single modality provides only a partial and potentially unreliable view of human emotion.
Video-based methods can be hindered when individuals mask their emotions~\citep{iwasaki2016hiding} or when visual cues are degraded by occlusion or adverse lighting~\citep{elsheikh2024improved}. Conversely, EEG provides access to internal neural states but remains sensitive to inter-subject variability and sensor artifacts~\citep{peng2022joint}.
These complementary properties motivate their joint modeling.
Nevertheless, only a few studies have directly explored EEG-video emotion recognition~\citep{lee2024eav, wang2026milmer}, and they mainly fuse globally aggregated features or predictions without explicitly modeling EEG anatomy and video temporal structure.
Although recent EEG foundation models provide strong global representation, three challenges remain.
First, global pooling discards channel-level anatomical organization.
Second, video is often represented by a single aggregated feature, obscuring the temporal evolution of facial behavior.
Third, existing EEG-video fusion methods lack a unified mechanism for relating these fine-grained structures to pretrained global representations.

To address these challenges, we introduce a novel EEG-Video Emotion Recognition (EVER) framework that combines strong pretrained backbones with graph-based structural modeling and inter-modal emotion reasoning.
For EEG, we introduce a Brain Anatomy-aware Regional Graph Convolutional Network (BAR-GCN). BAR-GCN groups EEG channels into anatomically meaningful regions and models their relationships to construct region-level representations.
For video, we introduce a Video Clip Gaussian temporal prior (VCG), which retains individual clip features as temporally ordered nodes. Sinusoidal positional encoding provides clip-order information, while learnable Gaussian weighting regulates clip-to-clip affinities according to temporal distance.
Finally, EVER constructs an inter-modal graph over the global EEG embedding, montage-flexible EEG region nodes, and temporally ordered video clip nodes.
Unlike conventional EEG-video methods that operate on globally aggregated modality representations, EVER jointly preserves and refines global and fine-grained structural information within a unified graph.

To systematically evaluate our framework, we establish an extensive benchmark on two public datasets including Emognition~\citep{saganowski2022emognition} and MDMER~\citep{yang2024multimodal}.
Both datasets provide time-synchronized EEG and video recordings, enabling the joint analysis of observable behavioral cues and neural responses.
We enforce subject-independent splits to prevent participant overlap between training and evaluation sets and to evaluate performance on unseen subjects.
Our benchmark covers unimodal and multi-modal settings with 12 representative models, including five general video transformers, five EEG-specific foundation and transformer models, and two EEG-video fusion architectures.
Given the limited scale of paired EEG-video datasets, we employ pretrained modality-specific backbones and focus on structural modeling and inter-modal fusion.
Extensive experiments show that EVER achieves the highest average accuracy among the evaluated methods on both datasets.

\section{Related Work}
\label{sec:Related Work}
\subsection{Video and EEG Emotion Recognition}
\label{subsec:Video and EEG Emotion Recognition}
Video- and EEG-based emotion recognition have been widely studied as unimodal tasks. Existing video approaches often employ task-specific architectures tailored to facial expressions or affective behavior~\citep{zhang2019learning, zhang2021real, ben2021video}, while EEG methods typically design specialized sequence or graph networks to capture temporal dependencies and inter-channel relationships~\citep{gao2022eeg, du2022multi, pan2023msfr}.
Although effective within individual datasets, these models are commonly optimized for individual datasets and often provide limited transferability across recording conditions or acquisition setups.

These limitations motivate the use of pretrained foundation backbones. Large-scale pretraining provides more transferable representations and is particularly valuable for EEG-video emotion recognition, where paired multi-modal datasets are relatively small.
Accordingly, we benchmark representative video transformers, including ViViT~\citep{arnab2021vivit}, TimeSformer~\citep{bertasius2021space}, Video Swin Transformer~\revb{\citep{liu2022video}}, VideoMAE~\citep{tong2022videomae}, and AdaMAE~\citep{bandara2023adamae}. We also evaluate EEG-specific foundation models, including BIOT~\citep{yang2023biot}, LaBraM~\citep{jiang2024labram}, CBraMod~\citep{wang2025cbramod}, and REVE~\citep{el2026reve}, together with ST-Transformer~\citep{song2021sttransformer}. These models provide strong modality-specific representations and enable a more reliable comparison of unimodal performance.
Despite advances in pretrained representation learning, EEG-video fusion remains underexplored. Existing methods generally combine globally pooled modality features or aggregate instance-level predictions~\citep{lee2024eav, wang2026milmer}. These methods neglect the anatomical organization of EEG channels and the temporal evolution of facial behavior.

To address these limitations, EVER augments general pretrained backbones with structured graph reasoning. BAR-GCN constructs anatomically informed regional EEG representations, the Video Clip Gaussian temporal prior (VCG) regulates relationships among temporally ordered video clips, and an inter-modal graph integrates these structured nodes with the global EEG embedding.

\subsection{Graph Convolutional Networks}
\label{subsec:Graph Convolutional Networks}

Graph Convolutional Networks (GCNs)~\citep{kipf2016semi} have been widely adopted in EEG emotion recognition to model dependencies among electrodes. Existing methods typically construct graphs based on spatial distance, functional connectivity, or anatomical priors~\citep{gao2022eeg, du2022multi, pan2023msfr}.
Several studies further model EEG signals at multiple spatial scales. HD-GCN~\citep{ye2022hierarchical} captures local and global dependencies through separate graph branches. PGCN~\citep{jin2024pgcn} aggregates electrode features at local, regional, and global levels, while \revb{MD-GCN}~\citep{du2022multi} models interactions among channels and predefined regions.
Although these methods exploit regional EEG structure, their graph definitions are often tied to the electrode layout, spatial coordinates, or regional partition of a specific montage. Applying them to datasets with different channel configurations therefore requires reconstructing the graph or redefining its regional organization. In addition, their reasoning remains confined to EEG and does not explicitly connect regional neural activity with another modality.

Graph reasoning has also been explored in video understanding by treating objects or temporal segments as nodes~\citep{wang2018videos, yan2018spatial, yang2020spatial}. These methods commonly rely on appearance similarity, local temporal connections, or predefined relations. Learnable Gaussian kernels have also been introduced for long-range temporal modeling~\citep{piergiovanni2019temporal}, but primarily in the form of temporal convolution rather than graph interaction among pretrained clip features.
Consequently, existing EEG and video graph methods remain modality-specific and do not provide a unified structure for multi-modal emotion reasoning.

To address these limitations, EVER introduces montage-flexible anatomical modeling and graph reasoning over pretrained modality representations.
BAR-GCN maps electrodes from heterogeneous montages into five canonical scalp regions defined by the international 10-20 system~\revb{\citep{klem1999ten}}, producing region nodes without dataset-specific graph reconstruction.
The VCG temporal prior converts pretrained video features into temporally ordered clip nodes and modulates their affinities using positional encoding and learnable Gaussian distance weights.
These fine-grained EEG and video nodes are jointly refined with the global EEG foundation-model representation through a unified inter-modal graph.
Unlike prior methods that either perform modality-specific graph reasoning or fuse globally pooled features, EVER directly connects pretrained global representations with anatomy-aware EEG regions and temporally structured video clips.

\section{Method}
\label{sec:Method}

\begin{figure*}[t!]
    \centering
    \includegraphics[width=0.99\linewidth]{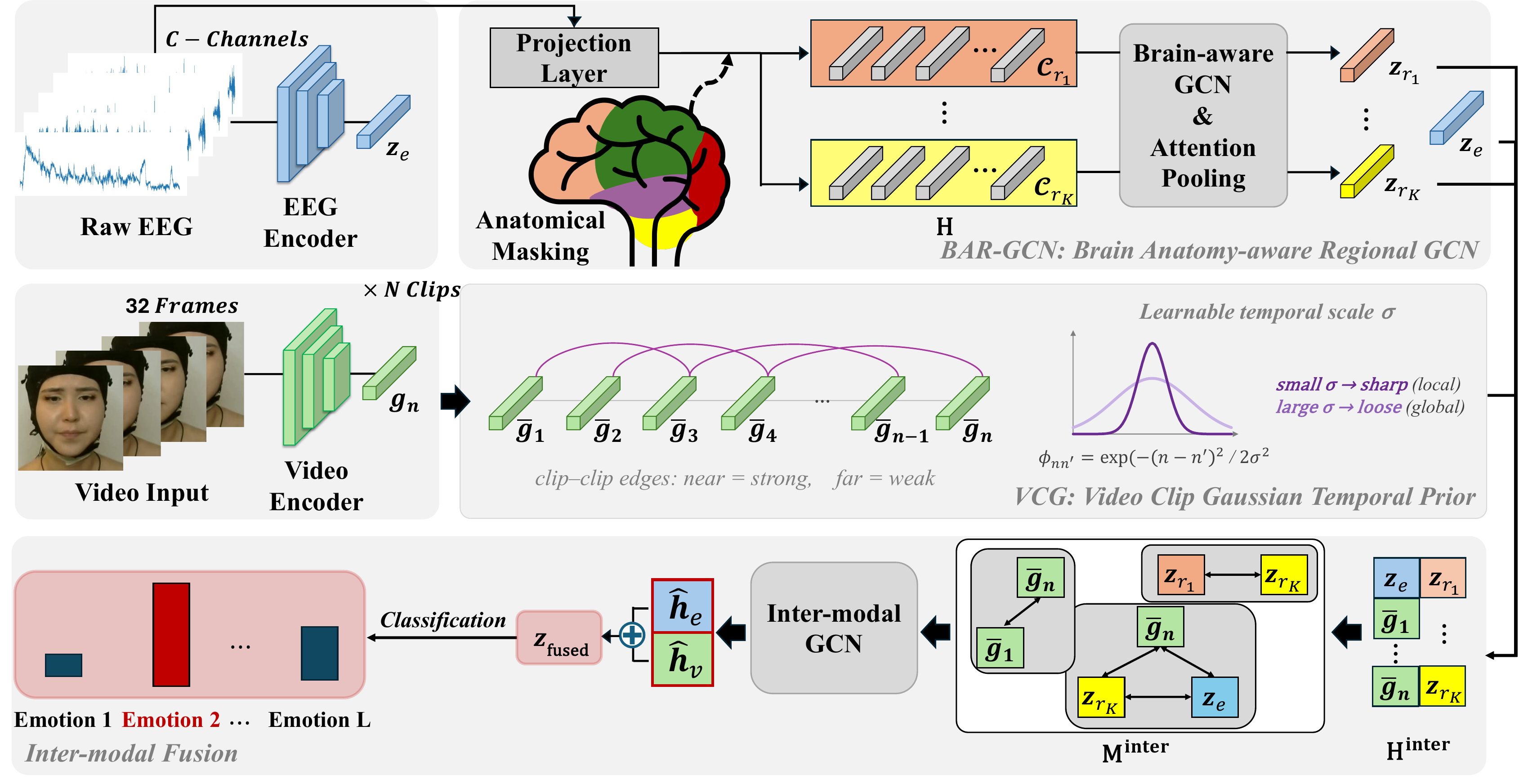}
    \caption{\textbf{Overall framework of the EEG-Video Emotion Recognition (EVER).} The global EEG embedding $\mathbf{z}_e$ and the temporally ordered video clip embeddings $\bar{\mathbf{g}}_{n}$ are extracted from synchronized raw EEG and video.
    In BAR-GCN, the per-channel EEG projection yields local nodes $\mathbf{H}$, which perform BAR-GCN with anatomical masking within region $\mathcal{C}_{r_k}$ and attention pooling to produce region features $\mathbf{z}_{r_k}$.
    The VCG temporal prior represents the video as $N$ temporally ordered clip nodes $\{\bar{\mathbf{g}}_{n}\}_{n=1}^{N}$ and modulates their affinities using the learnable Gaussian weight $\phi_{nn'}$.
    In the inter-modal fusion stage, we form $\mathbf{H}^{\text{inter}}$ and apply a mask $\mathbf{M}^{\text{inter}}$. Finally, emotion is predicted from the fused logit $\mathbf{z}_{\text{fused}}$ derived from $\hat{\mathbf{h}}_e$ and $\hat{\mathbf{h}}_v$.
    }
    \label{fig:Overall_Architecture}
\end{figure*}

\subsection{Overall Framework}
\label{subsec:Overall Framework}
\Figref{fig:Overall_Architecture} illustrates our EVER framework.
In detail, it consists of the following three components.

\subsubsection{Video Network}

We adopt MAE-based video backbones~\citep{tong2022videomae, bandara2023adamae} to extract visual representations.
Given a video $\mathbf{V}$, we divide its full temporal sequence into $N$ ordered clips $\{\mathbf{V}_{n}\}_{n=1}^{N}$, where $N$ varies with the video duration.
Each clip is uniformly sampled into 32 frames and represented as
$\mathbf{V}_{n}\in\mathbb{R}^{32\times3\times H\times W}$,
where $H$ and $W$ denote the frame height and width.
We retain the pretrained encoder $f_v$ and discard its decoder. All encoder blocks are frozen except the final block, which is fine-tuned jointly with EVER.
The shared encoder extracts each clip representation as:
\begin{equation}
\mathbf{g}_{n}=f_v(\mathbf{V}_{n})\in\mathbb{R}^{d},
\label{eq:video_network}
\end{equation}
where $d=768$ is the embedding dimension. This yields variable-length ordered clip embeddings.

\subsubsection{EEG Network}
Let the EEG input paired with the corresponding video be
$\mathbf{E}\in\mathbb{R}^{C\times T}$,
where $C$ and $T$ denote the number of channels and temporal samples, respectively.
The complete EEG sequence is resampled to 2,000 samples, globally z-score normalized, and divided into ten 200-sample patches for the pretrained CBraMod~\citep{wang2025cbramod} encoder as follows:
\begin{equation}
\mathbf{z}_e=f_e(\mathbf{E})\in\mathbb{R}^{d}.
\label{eq:eeg_network}
\end{equation}
In parallel, the same raw EEG signal is processed through a separate channel-wise projection pathway for BAR-GCN\revb{, which computes the short-time Fourier transform of each channel at its native sampling rate and thereby preserves the fine spectral content}.
This dual-path design captures global temporal dynamics through the pretrained encoder while preserving channel-level anatomical structure through BAR-GCN.
Since each label describes the complete stimulus interval rather than individual clips, we use a global EEG node while preserving local temporal variation through the video clip nodes.
\revb{The two modalities are related at the trial level, and the temporal structure of the video interacts with the EEG through the inter-modal graph, in which every clip node is connected to the global EEG node and to all region nodes.}

\subsubsection{Fusion Module}
Given the unimodal representations, EVER performs fusion through a unified inter-modal graph.
BAR-GCN organizes channel-level EEG features into anatomically grounded region nodes, while the VCG temporal prior equips the video clip nodes with explicit temporal order and distance information.
These structured nodes, together with the global EEG embedding, form the inter-modal graph. GCN propagates information across the two modalities, and the refined EEG and video node representations are used to produce the emotion logits.

\subsection{Brain Anatomy-aware Regional GCN}
\label{subsec:Brain Anatomy-aware Regional GCN}
The raw EEG signal $\mathbf{E}\in\mathbb{R}^{C\times T}$ is processed independently of the global foundation-model pathway.
For each channel, we compute the log-magnitude STFT using a Hann window, with $n_{\mathrm{fft}}=512$ for Emognition and $128$ for MDMER and a hop size of $n_{\mathrm{fft}}/4$.
The resulting spectrum $\tilde{\mathbf{E}}$ is resized to $128\times256$, flattened, and projected to dimension $d$ as:
\begin{equation}
\label{eq:EEG_node}
\mathbf{H}
=
\proj\left(\flatten(\tilde{\mathbf{E}})\right)
\in\mathbb{R}^{C\times d},
\end{equation}
where $\mathbf{H}$ is the channel-level node matrix and
$\mathbf{h}_i\in\mathbb{R}^{d}$ denotes the embedding of the $i$-th channel.
This projection preserves the full frequency-time input of each channel without temporal pooling, while compressing it into the common graph dimension $d$.

Motivated by evidence that brain anatomy emphasizes region-specific interactions, we design a region-level graph that embeds anatomical priors and enables the network to capture intra-region relationships through GCN.
Let $\mathcal{R} = \{{r_{k}}\}_{k=1}^{K}
$ denote $K$ predefined brain regions.
Each region $r_k$ is associated with a subset of EEG channels, represented by the index set $\mathcal{C}_{r_k} \subseteq \{{1,\dots,C}\}$.
Based on this anatomical prior, the local adjacency matrix $\tilde{\mathbf{A}}_{ij}$ is computed by cosine similarity with region-wise masking as follows:
\begin{equation}
\label{eq:local_adjacency}
\tilde{\mathbf{A}}_{ij} =
\frac{\mathbf{h}_i^{\top}\mathbf{h}_j}{\|\mathbf{h}_i\|\,\|\mathbf{h}_j\|}
\cdot \mathbf{M}^{\text{local}}_{ij},\
\quad i,j \in [1, C].
\mathbf{M}^{\text{local}}_{ij}
=
\begin{cases}
1, &
\text{if } \exists k \ \text{s.t. }
i,j \in \mathcal{C}_{r_k}
\text{ and } i \neq j,\\
0, & \text{otherwise},
\end{cases}
\end{equation}
where $\mathbf{M}^{\text{local}}\in\{0,1\}^{C\times C}$ encodes whether channels $i$ and $j$ belong to the same brain region $r_k$.
We retain signed cosine similarities without rectification, allowing dissimilar nodes to exert subtractive influence.
After masking and symmetrization, the degree is computed from the signed row sum and lower-bounded by $\epsilon=10^{-3}$ before normalization.
We first obtain the symmetrically normalized adjacency matrix and then apply a two-layer GCN over all EEG channels as follows:
\begin{equation}
\label{eq:symmetric}
\begin{aligned}
\hat{\mathbf{A}}
=
\mathbf{D}^{-1/2}\tilde{\mathbf{A}}\mathbf{D}^{-1/2}, \
\
\mathbf{H}^{\text{local}} =
\operatorname{GCN}_{2}
\left(
\operatorname{GELU}
\left(
\operatorname{GCN}_{1}
(\mathbf{H},\hat{\mathbf{A}})
\right),
\hat{\mathbf{A}}
\right),
\end{aligned}
\end{equation}
where $\mathbf{D}$ denotes the degree matrix of the symmetrized adjacency $\tilde{\mathbf{A}}$, and $\hat{\mathbf{A}}$ is its symmetrically normalized form. $\mathbf{H}$ denotes the input channel representations, while $\mathbf{H}^{\text{local}}$ represents the region-aware channel features obtained by two GCN layers~\citep{kipf2016semi} with a GELU activation between them.

To obtain compact region-level representations, we group the channel outputs according to the anatomical partition $\mathcal{R}$ and apply attention pooling within each brain region $r_k$.
Specifically, attention pooling assigns a learnable weight $\alpha_{k,i}$ to each channel representation $\mathbf{h}^{\text{local}}_i$ within a region using a parameter vector \revb{$\mathbf{w}_a$}, and aggregates them into a region-level embedding \revb{$\mathbf{z}_{r_k} \in \mathbb{R}^d$} as:

\begin{equation}
\label{eq:attention_pooling}
\alpha_{k,i}
=
\frac{
\exp\!\left(
\mathbf{w}_{a}^{\top}
\mathbf{h}^{\text{local}}_i
\right)
}{
\sum\limits_{j\in\mathcal{C}_{r_k}}
\exp\!\left(
\mathbf{w}_{a}^{\top}
\mathbf{h}^{\text{local}}_j
\right)
},
\
\
\mathbf{z}_{r_k}
=
\sum_{i\in\mathcal{C}_{r_k}}
\alpha_{k,i}
\mathbf{h}^{\text{local}}_i.
\end{equation}

\subsection{Video Clip Gaussian Temporal Prior}
\label{subsec:Video Clip Gaussian Temporal Prior}
While the EEG branch models fine-grained channel interactions guided by anatomical priors, encoding the whole video into a single embedding collapses the entire sequence into one vector and discards the temporal evolution of facial behavior, which is a salient cue for emotion.

To preserve this temporal structure, we introduce a Video Clip Gaussian temporal prior (VCG). It retains individual clip embeddings as temporally ordered nodes and uses learnable Gaussian weighting to regulate their affinities in the subsequent inter-modal graph.
Specifically, we uniformly partition the video along the temporal axis into $N$ densely sampled clips $\{\mathbf{V}_{n}\}_{n=1}^{N}$, and encode each clip with the shared video backbone $f_{v}$ as defined in \eqnref{eq:video_network}.
This yields the clip node matrix $\mathbf{G} = [\mathbf{g}_{1}, \dots, \mathbf{g}_{N}]^{\top} \in \mathbb{R}^{N \times d}$.
To make the graph aware of the temporal order of clips, we add a sinusoidal positional encoding $\mathbf{p}_{n} \in \mathbb{R}^{d}$ to each clip node as follows:
\begin{equation}
\label{eq:clip_pe}
\bar{\mathbf{g}}_{n} = \mathbf{g}_{n} + \mathbf{p}_{n}, \quad
\bar{\mathbf{G}} = [\bar{\mathbf{g}}_{1}, \dots, \bar{\mathbf{g}}_{N}]^{\top},
\end{equation}
where $\mathbf{p}_{n}$ is the standard transformer sinusoidal encoding of the clip index $n$.
Since affective cues in video are largely local in time, temporally adjacent clips are more likely to share the same emotional state.
We encode this inductive bias with a Gaussian temporal kernel that down-weights edges between distant clips as follows:
\begin{equation}
\label{eq:gaussian_kernel}
\phi_{nn'} = \exp\!\left(-\frac{(n-n')^{2}}{2\sigma^{2}}\right), \quad n,n' \in [1, N].
\end{equation}
Here, $\sigma$ is a single learnable bandwidth shared across samples and reused by the inter-modal GCN layers, providing a lightweight dataset-level temporal prior.
It is optimized in log space, initialized to $3.0$ in clip-index units.
Combining content similarity with the temporal prior, we define the Gaussian-modulated clip-to-clip affinity as follows:
\begin{equation}
\label{eq:clip_adjacency}
\tilde{\mathbf{A}}^{\text{clip}}_{nn'}
=
\frac{
\bar{\mathbf{g}}_{n}^{\top}\bar{\mathbf{g}}_{n'}
}{
\|\bar{\mathbf{g}}_{n}\|\,\|\bar{\mathbf{g}}_{n'}\|
}
\cdot
\phi_{nn'},
\quad n,n'\in[1,N].
\end{equation}
Here, the diagonal entries are set to zero to exclude self-loops.
Unlike a single pooled embedding, the temporally ordered clip nodes $\bar{\mathbf{G}}$ preserve the evolution of facial behavior throughout the video, while $\phi_{nn'}$ regulates their affinities according to temporal distance.
Rather than propagating a separate video-only graph, we use $\tilde{\mathbf{A}}^{\text{clip}}$ as the clip-to-clip block of the inter-modal graph, so that the clip nodes are jointly refined with the global EEG and brain-region nodes.

\subsection{Inter-modal Fusion}
\label{subsec:Inter-modal Fusion}

In the final stage, we construct a unified inter-modal graph over the ordered video clip nodes $\{ \bar{\mathbf{g}}_{n} \}_{n=1}^{N}$, the global EEG embedding $\mathbf{z}_{e}$, and the brain-region features $\{ \mathbf{z}_{r_{k}} \}_{k=1}^{K}$.
The clip and global EEG representations serve as modality nodes, while the region features serve as local nodes.
The stacked node matrix is defined as follows:
\begin{equation}
\label{eq:all_node_set}
\mathbf{H}^{\text{inter}}
=
[\bar{\mathbf{g}}_{1}, \dots, \bar{\mathbf{g}}_{N},
\mathbf{z}_{e},
\mathbf{z}_{r_1}, \dots, \mathbf{z}_{r_K}]^\top
.
\end{equation}

Instead of a fully connected or purely similarity-driven graph, we inject domain knowledge through a structured binary mask $\mathbf{M}^{\text{inter}}$ that retains only semantically valid connections, namely clip-to-clip edges for temporal dynamics, clip-to-EEG edges for cross-modal interaction, modality-to-region edges bridging global and local information, and inter-region edges reflecting intra-brain dependencies as follows:
\begin{equation}
\label{eq:global_adjacency}
\mathbf{M}^{\text{inter}}_{ij}
=
\left\{
\begin{array}{ll}
1,
&
\begin{aligned}
\text{if } (i,j)\in\{&
(\bar{\mathbf{g}}_{n},\bar{\mathbf{g}}_{n'}),
(\bar{\mathbf{g}}_{n},\mathbf{z}_{e}),\\
&
(\bar{\mathbf{g}}_{n},\mathbf{z}_{r_k}),
(\mathbf{z}_{e},\mathbf{z}_{r_k}),
(\mathbf{z}_{r_k},\mathbf{z}_{r_{k'}})
\},
\end{aligned}
\\[1mm]
0, & \text{otherwise}.
\end{array}
\right.
\end{equation}
where $n \neq n'$, and $r_k, r_{k'} \in \mathcal{R}$ with $k \neq k'$.
Following \eqnref{eq:local_adjacency}, affinities are computed as cosine similarities so that feature magnitude does not bias comparisons across clip, region, and global EEG nodes, where the clip-to-clip block is given by the Gaussian-modulated $\tilde{\mathbf{A}}^{\text{clip}}$ in \eqnref{eq:clip_adjacency}.
The resulting $\tilde{\mathbf{A}}^{\text{inter}}$ is restricted by $\mathbf{M}^{\text{inter}}$ and symmetrically normalized following \eqnref{eq:symmetric} to obtain $\hat{\mathbf{A}}^{\text{inter}}$.

A two-layer GCN then integrates the temporally ordered clips, global EEG dynamics, and localized brain-region features into a unified representation, where the first layer preserves the embedding dimension and the second projects each node into the label space as follows:

\begin{equation}
\label{eq:global_gcn}
\hat{\mathbf{H}}
=
\operatorname{GCN}
\big(\mathbf{H}^{\text{inter}}, \hat{\mathbf{A}}^{\text{inter}}\big)\in \mathbb{R}^{(N+1+K) \times L},
\end{equation}
where $L$ denotes the number of target emotion classes.
Among the updated nodes, $\hat{\mathbf{h}}_{e}$ is the transformed embedding of $\mathbf{z}_e$, while the video representation is obtained by averaging the updated clip nodes. The final logit is then computed as a gated combination of these two representations as:
\begin{equation}
\label{eq:video_readout}
\mathbf{z}_{\text{fused}} = \boldsymbol{\omega} \odot \hat{\mathbf{h}}_{v}
+ (\mathbf{1} - \boldsymbol{\omega}) \odot \hat{\mathbf{h}}_{e}
\in \mathbb{R}^L, \
\
\hat{\mathbf{h}}_{v} = \frac{1}{N} \sum_{n=1}^{N} \hat{\mathbf{g}}_{n} \in \mathbb{R}^L
\end{equation}
where $\boldsymbol{\omega}=\operatorname{sigmoid}(\mathbf{w}_{f})$ is a global per-class fusion gate shared across samples.
This lightweight gate learns class-specific modality preferences while avoiding a sample-dependent gating network.

\subsection{Loss Functions}
\label{subsec:Loss Functions}
To ensure consistent comparison, we adopt the following unified loss for all benchmark models as:
\begin{equation}
\label{eq:loss_base}
\mathcal{L}_{\text{total}} =
\lambda_{\text{ce}}\mathcal{L}_{\text{ce}} + \lambda_{\text{focal}}\mathcal{L}_{\text{focal}},
\end{equation}
where $\mathcal{L}_{\text{ce}}$ denotes the weighted cross-entropy loss~\citep{cui2019class} to mitigate class imbalance arising from participant's self-assessments, and $\mathcal{L}_{\text{focal}}$ denotes the focal loss to further address hard-to-classify samples.

\section{Experimental Results}
\label{sec:Experimental Results}

\subsection{Implementation Details}
\label{subsec:Implementation Details}
\subsubsection{Datasets}
\label{subsubsec:Datasets}

We conduct experiments on two public datasets including MDMER~\citep{yang2024multimodal} and Emognition~\citep{saganowski2022emognition}, for which we collect paired EEG–video samples.
\textbf{MDMER} dataset involves recordings from 73 participants across 32 emotion-eliciting video clips. From this dataset, we collect 2.3K paired EEG–video samples, of which 1.8K (58 participants) are used for training and \revb{475} (15 participants) for evaluation.
EEG was recorded with 18 channels, and videos were captured at 30 fps with $640 \times 480$ resolution. Ground-truth annotations were obtained from self-assessments of valence, arousal, and dominance on a 9-point scale.
\textbf{Emognition} dataset contains recordings from 37 participants with affective multimedia stimuli. We construct \revb{404} paired EEG–video samples, which are further split into 319 samples (29 participants) for training and \revb{85} (8 participants) for evaluation.
EEG was recorded with 4 channels, and videos were acquired at either 60 fps (\revb{294} samples) or 30 fps (110 samples) with $1080 \times 1920$ resolution.

Self-assessments include valence, arousal, and motivation on a 9-point scale.
In our experiments, we evaluate valence and arousal recognition on both datasets.
The original 9-point ratings are discretized into five categories by treating the midpoint as neutral:
$\{1,2\}$, $\{3,4\}$, $\{5\}$, $\{6,7\}$, and $\{8,9\}$.
We map electrodes from both datasets to canonical scalp regions defined by the international 10-20 system.
Only regions containing available electrodes are instantiated. This yields five regions (\ie, frontal, temporal, central, parietal, and occipital) for MDMER and two regions (\ie, frontal and temporal) for Emognition.
Each sample uses the complete paired EEG and video interval. Videos are divided into variable-length sequences of 32-frame clips, with frames sampled at a temporal stride of four and consecutive clips overlapping by 50\%. The sampling stride is normalized to a 30 fps reference to maintain a consistent clip duration across frame rates.
Facial regions are cropped and resized to $224\times224$ following prior studies~\citep{zafeiriou2017aff, sun2024hicmae}.
The EEG encoder is fine-tuned end-to-end. 
Additional dataset details are provided in Appendix~\ref{app:Details of Dataset}, while the clip sampling configuration is analyzed in Appendix~\ref{app:Video Clip Configuration}.

\subsubsection{Environments}
\label{subsubsec:Environments}
All experiments use the same subject-independent splits, training budget, loss formulation, and evaluation metrics to ensure a fair comparison.
For the benchmark evaluation, all networks are trained for 100 epochs with a batch size of 12, while model-specific learning rates and weight decay follow their recommended configurations.
Existing multi-modal baselines retain their original architectures.
We report Accuracy, Unweighted Average Recall (UAR), and Weighted F1-score (W-F1).
We set $\lambda_{\text{ce}} = 0.9$ and $\lambda_{\text{focal}} = 0.1$ for all networks.
\revb{For every model, we report the checkpoint with the highest average accuracy on the evaluation split. Since the paired EEG–video datasets are small, we do not reserve validation subjects from the training split, and the same protocol is applied to all unimodal and multi-modal models for a fair comparison. To examine whether this selection accounts for the advantage of EVER, we additionally report an evaluation in which model selection is separated from the test subjects.}
More details are provided in the Appendix~\ref{app:Experimental Setup}.

\subsection{Benchmark Results}
\label{subsec:Benchmark Results}

\begin{table*}[t!]
\centering
\caption{\textbf{Performance comparison on the Emognition and MDMER datasets.} We report Valence, Arousal, and their average score in terms of Accuracy (Acc), UAR, and W-F1.}
\resizebox{0.82\linewidth}{!}{
\begin{tabular}{c|c|c|ccc|ccc}
\Xhline{3\arrayrulewidth}
\multirow{2}{*}{Modality} & \multirow{2}{*}{Model} 
& \multirow{2}{*}{Target} 
& \multicolumn{3}{c|}{\textbf{Emognition}} 
& \multicolumn{3}{c}{\textbf{MDMER}} \\ \cline{4-9}
& & & Acc & UAR & W-F1 & Acc & UAR & W-F1 \\ \hline\hline
\multirow{15}{*}{Video} 
& \multirow{3}{*}{ViViT~\citep{arnab2021vivit}} & Valence & \revb{0.188} & \revb{0.170} & \revb{0.180} & \revb{0.301} & 0.218 & 0.278 \\
& & Arousal & \revb{0.294} & \revb{0.226} & 0.248 & \revb{0.288} & \revb{0.229} & \revb{0.272} \\
& & \cellcolor{Gray1}Average & \cellcolor{Gray1}\revb{0.241} & \cellcolor{Gray1}\revb{0.198} & \cellcolor{Gray1}\revb{0.214} & \cellcolor{Gray1}0.295 & \cellcolor{Gray1}\revb{0.223} & \cellcolor{Gray1}\revb{\underline{0.275}} \\ \cline{2-9}
& \multirow{3}{*}{Swin~\revb{\citep{liu2022video}}} & Valence & \revb{0.294} & \revb{0.236} & \revb{0.256} & \revb{0.347} & \revb{0.245} & \revb{0.307} \\
& & Arousal & \revb{0.224} & \revb{0.186} & \revb{0.191} & \revb{0.293} & 0.201 & \revb{0.238} \\
& & \cellcolor{Gray1}Average & \cellcolor{Gray1}\revb{0.259} & \cellcolor{Gray1}\revb{0.211} & \cellcolor{Gray1}\revb{0.224} & \cellcolor{Gray1}0.320 & \cellcolor{Gray1}0.223 & \cellcolor{Gray1}\revb{0.272} \\ \cline{2-9}
& \multirow{3}{*}{TSF~\citep{bertasius2021space}} & Valence & \revb{0.294} & \revb{0.217} & \revb{0.218} & 0.309 & 0.202 & \revb{0.213} \\
& & Arousal & \revb{0.271} & \revb{0.238} & \revb{0.212} & \revb{0.305} & \revb{0.215} & \revb{0.235} \\
& & \cellcolor{Gray1}Average & \cellcolor{Gray1}\revb{0.282} & \cellcolor{Gray1}\revb{0.227} & \cellcolor{Gray1}\revb{0.215} & \cellcolor{Gray1}\revb{0.307} & \cellcolor{Gray1}\revb{0.208} & \cellcolor{Gray1}0.224 \\ \cline{2-9}
& \multirow{3}{*}{VideoMAE~\citep{tong2022videomae}} & Valence & \revb{0.282} & \revb{0.262} & \revb{0.275} & \revb{0.354} & \revb{0.228} & \revb{0.264} \\
& & Arousal & \revb{0.318} & \revb{0.268} & \revb{0.294} & \revb{0.314} & 0.211 & 0.178 \\
& & \cellcolor{Gray1}Average & \cellcolor{Gray1}\revb{0.300} & \cellcolor{Gray1}\revb{0.265} & \cellcolor{Gray1}\revb{0.284} & \cellcolor{Gray1}\revb{0.334} & \cellcolor{Gray1}\revb{0.219} & \cellcolor{Gray1}\revb{0.221} \\ \cline{2-9}
& \multirow{3}{*}{AdaMAE~\citep{bandara2023adamae}} & Valence & \revb{0.318} & \revb{0.235} & \revb{0.224} & \revb{0.358} & \revb{0.232} & \revb{0.284} \\
& & Arousal & \revb{0.306} & \revb{0.275} & \revb{0.238} & 0.305 & 0.218 & 0.235 \\
& & \cellcolor{Gray1}Average & \cellcolor{Gray1}\revb{0.312} & \cellcolor{Gray1}\revb{0.255} & \cellcolor{Gray1}\revb{0.231} & \cellcolor{Gray1}0.332 & \cellcolor{Gray1}0.225 & \cellcolor{Gray1}0.260 \\ \hline
\multirow{15}{*}{EEG} 
& \multirow{3}{*}{ST-Transformer~\citep{song2021sttransformer}} & Valence & 0.235 & 0.214 & 0.187 & 0.236 & 0.174 & 0.151 \\
& & Arousal & 0.400 & 0.320 & 0.332 & 0.265 & 0.196 & 0.251 \\
& & \cellcolor{Gray1}Average & \cellcolor{Gray1}0.318 & \cellcolor{Gray1}0.267 & \cellcolor{Gray1}0.259 & \cellcolor{Gray1}0.251 & \cellcolor{Gray1}0.185 & \cellcolor{Gray1}0.201 \\ \cline{2-9}
& \multirow{3}{*}{BIOT~\citep{yang2023biot}} & Valence & 0.341 & 0.334 & 0.329 & 0.248 & 0.203 & 0.243 \\
& & Arousal & 0.306 & 0.236 & 0.283 & 0.255 & 0.221 & 0.258 \\
& & \cellcolor{Gray1}Average & \cellcolor{Gray1}0.324 & \cellcolor{Gray1}0.285 & \cellcolor{Gray1}\revb{\underline{0.306}} & \cellcolor{Gray1}0.252 & \cellcolor{Gray1}0.212 & \cellcolor{Gray1}0.251 \\ \cline{2-9}

& \multirow{3}{*}{LaBraM~\citep{jiang2024labram}} & Valence & 0.259 & 0.200 & 0.106 & 0.309 & 0.189 & 0.172 \\
& & Arousal & 0.294 & 0.273 & 0.171 & 0.288 & 0.200 & 0.148 \\
& & \cellcolor{Gray1}Average & \cellcolor{Gray1}0.276 & \cellcolor{Gray1}0.237 & \cellcolor{Gray1}0.139 & \cellcolor{Gray1}0.299 & \cellcolor{Gray1}0.195 & \cellcolor{Gray1}0.160 \\ \cline{2-9}
& \multirow{3}{*}{CBraMod~\citep{wang2025cbramod}} & Valence & 0.365 & 0.342 & 0.366 & 0.320 & 0.211 & 0.232 \\
& & Arousal & 0.341 & 0.232 & 0.218 & 0.284 & 0.206 & 0.206 \\
& & \cellcolor{Gray1}Average & \cellcolor{Gray1}\revb{\underline{0.353}} & \cellcolor{Gray1}0.287 & \cellcolor{Gray1}0.292 & \cellcolor{Gray1}0.302 & \cellcolor{Gray1}0.208 & \cellcolor{Gray1}0.219 \\ \cline{2-9}
& \multirow{3}{*}{REVE~\citep{el2026reve}} & Valence & 0.294 & 0.243 & 0.263 & 0.284 & 0.189 & 0.213 \\
& & Arousal & 0.259 & 0.233 & 0.218 & 0.282 & 0.220 & 0.256 \\
& & \cellcolor{Gray1}Average & \cellcolor{Gray1}0.276 & \cellcolor{Gray1}0.238 & \cellcolor{Gray1}0.240 & \cellcolor{Gray1}0.283 & \cellcolor{Gray1}0.205 & \cellcolor{Gray1}0.234 \\ \hline
\multirow{12}{*}{Video+EEG} 
& \multirow{3}{*}{EAV~\citep{lee2024eav}} & Valence & \revb{0.282} & 0.275 & \revb{0.252} & \revb{0.312} & 0.190 & \revb{0.189} \\
& & Arousal & \revb{0.259} & \revb{0.244} & \revb{0.155} & \revb{0.356} & 0.263 & \revb{0.273} \\
& & \cellcolor{Gray1}Average & \cellcolor{Gray1}\revb{0.271} & \cellcolor{Gray1}0.260 & \cellcolor{Gray1}\revb{0.203} & \cellcolor{Gray1}\revb{0.334} & \cellcolor{Gray1}\revb{0.227} & \cellcolor{Gray1}\revb{0.231} \\ \cline{2-9}
& \multirow{3}{*}{Milmer~\citep{wang2026milmer}} & Valence & \revb{0.224} & \revb{0.230} & \revb{0.163} & \revb{0.339} & 0.218 & 0.222 \\
& & Arousal & \revb{0.376} & \revb{0.281} & \revb{0.278} & \revb{0.297} & 0.207 & \revb{0.225} \\
& & \cellcolor{Gray1}Average & \cellcolor{Gray1}\revb{0.300} & \cellcolor{Gray1}\revb{0.256} & \cellcolor{Gray1}\revb{0.221} & \cellcolor{Gray1}\revb{0.318} & \cellcolor{Gray1}\revb{0.213} & \cellcolor{Gray1}\revb{0.224} \\ \cline{2-9}
& \multirow{3}{*}{\textbf{Ours (VideoMAE+CBraMod)}} & Valence & \revb{0.388} & \revb{0.373} & \revb{0.363} & \revb{0.377} & \revb{0.249} & \revb{0.313} \\
& & Arousal & \revb{0.353} & \revb{0.333} & \revb{0.343} & \revb{0.331} & 0.225 & \revb{0.248} \\
& & \cellcolor{Gray1}Average & \cellcolor{Gray1}\revb{\textbf{0.371}} & \cellcolor{Gray1}\textbf{0.353} & \cellcolor{Gray1}\revb{\textbf{0.353}} & \cellcolor{Gray1}\revb{\textbf{0.354}} & \cellcolor{Gray1}\revb{\textbf{0.237}} & \cellcolor{Gray1}\revb{\textbf{0.280}} \\ \cline{2-9}
& \multirow{3}{*}{\textbf{Ours (AdaMAE+CBraMod)}} & Valence & \revb{0.353} & \revb{0.306} & \revb{0.288} & \revb{0.347} & \revb{0.228} & \revb{0.284} \\
& & Arousal & \revb{0.388} & \revb{0.306} & \revb{0.308} & \revb{0.335} & \revb{0.233} & \revb{0.264} \\
& & \cellcolor{Gray1}Average & \cellcolor{Gray1}\revb{\textbf{0.371}} & \cellcolor{Gray1}\revb{\underline{0.306}} & \cellcolor{Gray1}\revb{0.298} & \cellcolor{Gray1}\revb{\underline{0.341}} & \cellcolor{Gray1}\revb{\underline{0.231}} & \cellcolor{Gray1}\revb{0.274} \\ 

\Xhline{3\arrayrulewidth}
\multicolumn{9}{r}{\textbf{Bold}: The best, \underline{Underline}: The second-best} \\
\end{tabular}}
\label{Tab:emognition_mdmer_benchmark}
\end{table*}

\Tabref{Tab:emognition_mdmer_benchmark} presents the results on Emognition and MDMER datasets under unimodal and multi-modal settings. Each baseline follows its training configuration to ensure a fair comparison.
The relative strength of the two modalities varies across datasets. On Emognition, the average accuracy of the EEG backbones ranges from 27.6\% to 35.3\%, exceeding that of the video backbones, which ranges from \revb{24.1\% to 31.2\%}. In contrast, video provides stronger representations on MDMER, achieving 29.5-33.4\% average accuracy compared with 25.1-30.2\% for EEG.
This dataset-dependent behavior indicates that neither modality is consistently dominant and motivates structured fusion that exploits their complementary information.
Existing multi-modal methods do not consistently improve upon the strongest unimodal models. EAV~\citep{lee2024eav} and Milmer~\citep{wang2026milmer} obtain average accuracies of \revb{27.1\% and 30.0\%} on the Emognition and \revb{33.4\% and 31.8\%} on the MDMER, respectively. These results \revb{do not exceed} CBraMod on the Emognition at 35.3\% and VideoMAE on MDMER at 33.4\%, showing that combining modalities does not guarantee improved performance.

In contrast, EVER \revb{achieves the highest average accuracy among the evaluated unimodal and multi-modal baselines on both datasets}.
\revb{A majority-class predictor obtains 30.0\% on the Emognition and 31.7\% on the MDMER (Appendix~\ref{app:Evaluation Reliability}). EVER exceeds this level by 7.1 and 3.7 percentage points.}
\revb{Both configurations achieve the highest average accuracy of 37.1\% on the Emognition, exceeding the strongest baseline by 1.8 percentage points.} On the MDMER, VideoMAE+CBraMod reaches \revb{35.4\%}, providing a \revb{2.0 percentage point} improvement over the best baseline.
VideoMAE+CBraMod also achieves the best balanced metrics, obtaining 35.3\% UAR and \revb{35.3\%} W-F1 on the Emognition. These scores improve upon the strongest baselines by 6.6 and \revb{4.7} percentage points, respectively. This indicates that the proposed fusion improves class-balanced recognition rather than relying primarily on dominant classes.
\revb{The two backbone configurations differ mainly in the class-balanced metrics.} VideoMAE+CBraMod performs best on UAR and W-F1, \revb{whereas the two configurations reach the same accuracy on the Emognition}. Nevertheless, both configurations \revb{achieve a higher average accuracy than the competing unimodal and multi-modal approaches}.
The gains arise from preserving anatomically structured EEG representations and temporally ordered video features, followed by node-level reasoning across the two modalities.
Additional analyses of statistical reliability and \revb{the unimodal and fused predictions}, together with comparisons against additional EEG architectures, are provided in Appendices~\ref{app:Bootstrap Confidence Intervals}, \ref{app:Complementarity}, and~\ref{app:Additional EEG Backbones}.

{
\subsection{Evaluation with Held-out Validation Subjects}
\label{subsec:Held-out Evaluation}

The results in \Tabref{Tab:emognition_mdmer_benchmark} are selected on the evaluation split.
To separate model selection from the final evaluation, we hold out eight validation subjects (256 trials) from the MDMER training split and use them for model selection.
The test subjects are the same as in \Tabref{Tab:emognition_mdmer_benchmark} and are used only for the final evaluation.
We conduct this experiment on MDMER since its larger number of subjects permits a validation set of this size.
The experiment is repeated over five seeds for EVER, its two unimodal backbones and their logit-averaging ensemble.

\begin{table}[t!]
\centering
\caption{\revb{\textbf{Evaluation with held-out validation subjects on the MDMER dataset.} Mean $\pm$ standard deviation over five seeds on the 475 test trials.}}
\footnotesize
\resizebox{0.8\linewidth}{!}{
\begin{tabular}{c|ccc}
\Xhline{3\arrayrulewidth}
Model & Acc & UAR & W-F1 \\ \hline\hline
CBraMod & 0.244 $\pm$ 0.021 & 0.206 $\pm$ 0.013 & 0.218 $\pm$ 0.016 \\
VideoMAE & 0.251 $\pm$ 0.015 & \underline{0.210 $\pm$ 0.009} & \underline{0.243 $\pm$ 0.013} \\
Logit-averaging ensemble (VideoMAE+CBraMod) & \underline{0.254 $\pm$ 0.018} & \underline{0.210 $\pm$ 0.016} & 0.226 $\pm$ 0.019 \\ \hline
\textbf{Ours (VideoMAE+CBraMod)} & \textbf{0.259 $\pm$ 0.012} & \textbf{0.215 $\pm$ 0.012} & \textbf{0.251 $\pm$ 0.007} \\
\Xhline{3\arrayrulewidth}
\multicolumn{4}{r}{\textbf{Bold}: The best, \underline{Underline}: The second-best} \\
\end{tabular}}
\label{Tab:heldout_mdmer}
\end{table}

As shown in \Tabref{Tab:heldout_mdmer}, EVER retains the highest mean in all three metrics.
This protocol is stricter than the one in \Tabref{Tab:emognition_mdmer_benchmark} and the scores of all models decrease.
A majority-class predictor obtains 0.317, 0.200, and 0.153 in accuracy, UAR, and W-F1, respectively.
The accuracies are below this level and the UAR values are close to it, which reflects the difficulty of the held-out setting.
In the class-balanced metrics, EVER improves the UAR over the compared models by 0.005 to 0.009 and the W-F1 by 0.008 to 0.033.
To examine the statistical significance, we conduct a paired bootstrap over the test trials.
EVER achieves a significantly higher W-F1 than CBraMod by 0.033 ($p < 0.001$) and than the ensemble by 0.025 ($p = 0.002$).
The differences in accuracy and UAR are not statistically significant, and the margin over VideoMAE lies within one standard deviation.
Nevertheless, the ordering of EVER and its two unimodal backbones remains consistent with \Tabref{Tab:emognition_mdmer_benchmark} across all three metrics.
The Emognition dataset contains 37 subjects, and only four validation subjects (44 trials) can be held out from its training split.
This number is too small for a stable selection, and the held-out evaluation is not conducted on Emognition.
}

\subsection{Ablation Study}
\label{subsec:Ablation Study}

We report the mean of valence and arousal scores. All fusion ablations use identical AdaMAE and CBraMod backbones to isolate the fusion mechanism.
\revb{All ablation studies are conducted under the evaluation protocol described in \secref{subsubsec:Environments}.
The components of EVER and the structure of BAR-GCN are analyzed on MDMER, which provides the larger test set. The VCG prior and the fusion strategy are analyzed on Emognition.}
Additional analyses of robustness to test-time EEG corruption and model size are provided in Appendices~\ref{app:Robustness to EEG Corruption} and~\ref{app:Model Size and Accuracy}.

\subsubsection{Component Analysis}
\label{subsubsec:Component Analysis}

\revb{\Tabref{Tab:ablation_components} evaluates the components of EVER on the MDMER dataset, where the unimodal video and EEG models achieve accuracies of 33.2\% and 30.2\%.
Introducing BAR-GCN region nodes into the inter-modal graph with a pooled video representation yields 28.1\%, indicating that the region-level EEG structure alone is insufficient.
Clip nodes retain clip embeddings in the inter-modal graph, whereas the VCG prior further equips them with positional encoding and Gaussian temporal modulation.
Introducing video clips as graph nodes without temporal structure increases the accuracy to 32.3\% but reduces the W-F1 to 17.1\%. This suggests that unordered clip representations introduce temporal ambiguity.
Incorporating the VCG prior increases the accuracy to 34.1\% and the W-F1 to 27.4\%, and the full model is the only configuration that exceeds both unimodal models in all three metrics.
The corresponding ablation on Emognition is reported in \Tabref{Tab:ablation_components_emog} of the appendix.}

\begin{table*}[t!]
\centering

\begin{minipage}[t]{0.55\linewidth}
\centering
\caption{\textbf{Ablation study of EVER components on the \revb{MDMER} dataset.}}
\footnotesize
\resizebox{\linewidth}{!}{
\begin{tabular}{c|ccc|ccc}
\Xhline{3\arrayrulewidth}
\multirow{2}{*}{Modality} 
& \multirow{2}{*}{BAR-GCN} 
& \multirow{2}{*}{Clip nodes} 
& \multirow{2}{*}{VCG}
& \multicolumn{3}{c}{\revb{\textbf{MDMER}}} \\ \cline{5-7}
& & & & Acc & UAR & W-F1 \\ \hline\hline
Video & & & & \revb{\underline{0.332}} & \revb{\underline{0.225}} & \revb{\underline{0.260}} \\ \hline
EEG & & & & \revb{0.302} & \revb{0.208} & \revb{0.219} \\ \hline
\multirow{3}{*}{Video + EEG}
& \checkmark & & & \revb{0.281} & \revb{0.188} & \revb{0.220} \\
& \checkmark & \checkmark & & \revb{0.323} & \revb{0.205} & \revb{0.171} \\
& \checkmark & \checkmark & \checkmark & \revb{\textbf{0.341}} & \revb{\textbf{0.231}} & \revb{\textbf{0.274}} \\
\Xhline{3\arrayrulewidth}
\multicolumn{7}{r}{\textbf{Bold}: The best, \underline{Underline}: The second-best} \\
\end{tabular}}
\label{Tab:ablation_components}
\end{minipage}
\hfill
\begin{minipage}[t]{0.44\linewidth}
\centering
\caption{\textbf{Ablation study of the BAR-GCN on the MDMER dataset.}}
\footnotesize
\resizebox{\linewidth}{!}{
\begin{tabular}{c|c|ccc}
\Xhline{3\arrayrulewidth}
\multirow{2}{*}{Method} & \multirow{2}{*}{Variant}
& \multicolumn{3}{c}{\textbf{MDMER}} \\ \cline{3-5}
& & Acc & UAR & W-F1 \\ \hline\hline
\multirow{5}{*}{Ours}
& Full & \revb{\textbf{0.341}} & \revb{\textbf{0.231}} & \revb{\textbf{0.274}} \\
& Random region partition & \underline{0.326} & \underline{0.208} & 0.185 \\
& w/o inter-region edges & \revb{0.319} & 0.205 & 0.196 \\
& w/o $\mathbf{M}^{\text{local}}$ & \revb{0.312} & 0.198 & \revb{0.155} \\
& Pooled region features & \revb{0.299} & 0.206 & \revb{\underline{0.207}} \\
\Xhline{3\arrayrulewidth}
\multicolumn{5}{r}{\textbf{Bold}: The best, \underline{Underline}: The second-best} \\
\end{tabular}}
\label{Tab:ablation_bargcn}
\end{minipage}
\end{table*}

\subsubsection{Effect of Anatomical Structure}
\label{subsubsec:Effect of Anatomical Structure}

\Tabref{Tab:ablation_bargcn} evaluates BAR-GCN on MDMER dataset, whose montage spans five anatomical regions.
Replacing the anatomical partition with a random channel-to-region partition of the same size retains an accuracy of 32.6\% but reduces W-F1 from \revb{27.4\%} to 18.5\%. This result suggests that grouping channels alone can preserve predictive performance, whereas anatomically meaningful assignments are important for balanced class recognition.
Removing the intra-region mask $\mathbf{M}^{\text{local}}$ and the inter-region edges further reduces the accuracy to \revb{31.2\% and 31.9\%}, respectively.
Pooling the region features before constructing the inter-modal graph produces the lowest accuracy of \revb{29.9\%}.

\subsubsection{Analysis of the VCG Prior}
\label{subsubsec:Analysis of the VCG Prior}

\begin{table*}[t!]
\centering
\begin{minipage}[t]{0.49\linewidth}
\centering
\caption{\textbf{Ablation of the VCG temporal prior on the Emognition dataset.} 
$\mathbf{p}_n$ denotes the positional encoding and $\phi_{nn'}$ denotes the learnable Gaussian prior.}
\footnotesize
\resizebox{\linewidth}{!}{
\begin{tabular}{c|c|ccc}
\Xhline{3\arrayrulewidth}
\multirow{2}{*}{Method} & \multirow{2}{*}{Variant}
& \multicolumn{3}{c}{\textbf{Emognition}} \\ \cline{3-5}
& & Acc & UAR & W-F1 \\ \hline\hline
\multirow{3}{*}{Ours (VCG prior)}
& Full & \revb{\textbf{0.371}} & \revb{\textbf{0.306}} & \revb{\textbf{0.298}} \\
& w/o $\mathbf{p}_n$ & \revb{\underline{0.294}} & \revb{\underline{0.257}} & \revb{\underline{0.231}} \\
& w/o $\phi_{nn'}$ & \revb{0.276} & \revb{0.246} & \revb{0.173} \\
\Xhline{3\arrayrulewidth}
\multicolumn{5}{r}{\textbf{Bold}: The best, \underline{Underline}: The second-best} \\
\end{tabular}}
\label{Tab:ablation_vcggcn}
\end{minipage}
\hfill
\begin{minipage}[t]{0.49\linewidth}
\centering
\caption{\textbf{Ablation study of the fusion strategy on the Emognition dataset.} 
``Pooled modality nodes'' averages the clip and region features before inter-modal fusion.}
\footnotesize
\resizebox{\linewidth}{!}{
\begin{tabular}{c|c|ccc}
\Xhline{3\arrayrulewidth}
\multirow{2}{*}{Fusion} & \multirow{2}{*}{Variant}
& \multicolumn{3}{c}{\textbf{Emognition}} \\ \cline{3-5}
& & Acc & UAR & W-F1 \\ \hline\hline
\multirow{4}{*}{w/o graph}
& Concatenation & \revb{0.306} & \revb{0.272} & \revb{0.271} \\
& Cross-attention & \revb{0.300} & \revb{0.265} & \revb{0.279} \\
& FiLM & \revb{0.300} & \revb{\underline{0.304}} & \revb{0.280} \\
& \revb{Logit-averaging ensemble} & \revb{\underline{0.353}} & \revb{0.288} & \revb{\underline{0.294}} \\ \hline
\multirow{2}{*}{GCN}
& Pooled modality nodes & \revb{0.288} & 0.200 & \revb{0.130} \\
& Ours (Inter-modal) & \revb{\textbf{0.371}} & \revb{\textbf{0.306}} & \revb{\textbf{0.298}} \\
\Xhline{3\arrayrulewidth}
\multicolumn{5}{r}{\textbf{Bold}: The best, \underline{Underline}: The second-best} \\
\end{tabular}}
\label{Tab:ablation_fusion}
\end{minipage}
\end{table*}

\Tabref{Tab:ablation_vcggcn} evaluates the positional encoding $\mathbf{p}_n$ and Gaussian temporal weight $\phi_{nn'}$ within the VCG prior.
Removing either component reduces the accuracy from \revb{37.1\% to 29.4\% and 27.6\%}, respectively. This indicates that the VCG prior benefits from combining both components rather than from either alone.
\revb{Both partial variants fall below the EEG-only model at 35.3\%. On the Emognition dataset, the EEG modality is considerably stronger than video, and clip nodes without the complete temporal prior dilute the EEG representation instead of complementing it. The same ablation on the MDMER dataset is reported in \Tabref{Tab:ablation_vcg_mdmer} of the appendix.}

\subsubsection{Fusion Strategy}
\label{subsubsec:Fusion Strategy}

\Tabref{Tab:ablation_fusion} provides a controlled comparison with identical backbones, fine-tuning settings, and training objectives.
\revb{Concatenation, Cross-attention, and FiLM achieve accuracies of 30.6\%, 30.0\%, and 30.0\%, respectively.} \revb{The logit-averaging ensemble of the two unimodal backbones reaches 35.3\%, the strongest result without a graph.}
Applying a GCN only to globally pooled modality nodes yields \revb{28.8\%}, whereas operating the graph over video clip and EEG region nodes increases the accuracy to \revb{37.1\%}.
This comparison shows that the main benefit arises from structured node-level interaction rather than the fusion operator alone.
\revb{The fusion operators and the pooled-node variant fall below the EEG-only model, which is consistent with the weaker video modality diluting the stronger EEG modality. The comparison is extended to the VideoMAE backbone and to MDMER in \Tabref{Tab:fusion_more} of the appendix. On the MDMER dataset, concatenation, cross-attention, and FiLM predict only the majority class under the shared hyperparameters and thus yield identical results, as shown in \Figref{fig:cm_fusion}.}

\section{Conclusion}
\label{sec:Conclusion}
In this paper, we proposed EVER for EEG--video emotion recognition.
EVER combines anatomy-aware EEG modeling with temporally structured video reasoning through an inter-modal graph.
We also conducted an extensive benchmark across two public datasets and twelve representative models.
Experimental results show that EVER \revb{achieves the highest average accuracy among the evaluated unimodal and multi-modal approaches, while the margin over the strongest video-only backbone is modest under held-out model selection}.
The variable-length clip design improves temporal coverage but increases computation in proportion to the video duration. Future work will explore efficient clip selection to enable evaluation over additional subject-independent splits.

\bibliography{iclr2027_conference}

@string{CVPR = "{Proc. of Computer Vision and Pattern Recognition (CVPR)}"}

@string{CVPRW = "{Proc. of Computer Vision and Pattern Recognition Workshops (CVPRW)}"}

@string{ECCV = "{Proc. of European Conf. on Computer Vision (ECCV)}"}

@string{ICCV = "{Proc. of Int'l Conf. on Computer Vision (ICCV)}"}

@string{ICML = "{Proc. of Int'l Conf. on Machine Learning (ICML)}"}

@string{TPAMI = "{IEEE Transactions on Pattern Analysis and Machine Intelligence (TPAMI)}"}

@string{NIPS = "{Proc. of  Advances in Neural Information Processing Systems (NeurIPS)}"}

@string{AAAI = "{Proc. of AAAI Conf. on Artificial Intelligence (AAAI)}"}

@string{ICLR = "{Proc. of  Int'l Conf. on Learning Representations (ICLR)}"}

@string{ACCESS = "{IEEE Access}"}

@string{Neural_Computing = "{Neural Computing and Applications}"}

@string{Trans_affec = "{IEEE Transactions on Affective Computing}"}

@string{SPL = "{IEEE Signal Processing Letters}"}

@article{aina2024hybrid,
  title={A hybrid learning-architecture for mental disorder detection using emotion recognition},
  author={Aina, Joseph and Akinniyi, Oluwatunmise and Rahman, Md Mahmudur and Odero-Marah, Valerie and Khalifa, Fahmi},
  journal=Access,
  volume={12},
  pages={91410--91425},
  year={2024},
  publisher={IEEE}
}

@article{chowdary2023deep,
  title={Deep learning-based facial emotion recognition for human--computer interaction applications},
  author={Chowdary, M Kalpana and Nguyen, Tu N and Hemanth, D Jude},
  journal=Neural_Computing,
  volume={35},
  number={32},
  pages={23311--23328},
  year={2023},
  publisher={Springer}
}

@article{cortinas2023toward,
  title={Toward explainable affective computing: A review},
  author={Corti{\~n}as-Lorenzo, Karina and Lacey, Gerard},
  journal={IEEE Transactions on Neural Networks and Learning Systems},
  volume={35},
  number={10},
  pages={13101--13121},
  year={2023},
  publisher={IEEE}
}

@article{zhang2019learning,
  title={Learning affective video features for facial expression recognition via hybrid deep learning},
  author={Zhang, Shiqing and Pan, Xianzhang and Cui, Yueli and Zhao, Xiaoming and Liu, Limei},
  journal=Access,
  volume={7},
  pages={32297--32304},
  year={2019},
  publisher={IEEE}
}

@article{zhang2021real,
  title={Real-time video emotion recognition based on reinforcement learning and domain knowledge},
  author={Zhang, Ke and Li, Yuanqing and Wang, Jingyu and Cambria, Erik and Li, Xuelong},
  journal={IEEE Transactions on Circuits and Systems for Video Technology},
  volume={32},
  number={3},
  pages={1034--1047},
  year={2021},
  publisher={IEEE}
}

@article{ben2021video,
  title={Video-based facial micro-expression analysis: A survey of datasets, features and algorithms},
  author={Ben, Xianye and Ren, Yi and Zhang, Junping and Wang, Su-Jing and Kpalma, Kidiyo and Meng, Weixiao and Liu, Yong-Jin},
  journal=TPAMI,
  volume={44},
  number={9},
  pages={5826--5846},
  year={2021},
  publisher={IEEE}
}

@article{iwasaki2016hiding,
  title={Hiding true emotions: micro-expressions in eyes retrospectively concealed by mouth movements},
  author={Iwasaki, Miho and Noguchi, Yasuki},
  journal={Scientific Reports},
  volume={6},
  number={1},
  pages={22049},
  year={2016},
  publisher={Nature Publishing Group UK London}
}

@article{elsheikh2024improved,
  title={Improved facial emotion recognition model based on a novel deep convolutional structure},
  author={Elsheikh, Reham A and Mohamed, MA and Abou-Taleb, Ahmed Mohamed and Ata, Mohamed Maher},
  journal={Scientific Reports},
  volume={14},
  number={1},
  pages={29050},
  year={2024},
  publisher={Nature Publishing Group UK London}
}

@article{peng2022joint,
  title={Joint EEG feature transfer and semisupervised cross-subject emotion recognition},
  author={Peng, Yong and Liu, Honggang and Kong, Wanzeng and Nie, Feiping and Lu, Bao-Liang and Cichocki, Andrzej},
  journal={IEEE Transactions on Industrial Informatics},
  volume={19},
  number={7},
  pages={8104--8115},
  year={2022},
  publisher={IEEE}
}

@inproceedings{liu2024vbh,
  title={Vbh-gnn: Variational bayesian heterogeneous graph neural networks for cross-subject emotion recognition},
  author={Liu, Chenyu and Zhou, Xinliang and Zhu, Zhengri and Zhai, Liming and Jia, Ziyu and Liu, Yang},
  booktitle=ICLR,
  year={2024}
}

@article{lee2024eav,
  title={EAV: EEG-audio-video dataset for emotion recognition in conversational contexts},
  author={Lee, Min-Ho and Shomanov, Adai and Begim, Balgyn and Kabidenova, Zhuldyz and Nyssanbay, Aruna and Yazici, Adnan and Lee, Seong-Whan},
  journal={Scientific Data},
  volume={11},
  number={1},
  pages={1026},
  year={2024},
  publisher={Nature Publishing Group UK London}
}

@article{koelstra2011deap,
  title={Deap: A database for emotion analysis; using physiological signals},
  author={Koelstra, Sander and Muhl, Christian and Soleymani, Mohammad and Lee, Jong-Seok and Yazdani, Ashkan and Ebrahimi, Touradj and Pun, Thierry and Nijholt, Anton and Patras, Ioannis},
  journal=Trans_affec,
  volume={3},
  number={1},
  pages={18--31},
  year={2011},
  publisher={IEEE}
}

@article{zheng2015investigating,
  title={Investigating critical frequency bands and channels for EEG-based emotion recognition with deep neural networks},
  author={Zheng, Wei-Long and Lu, Bao-Liang},
  journal={IEEE Transactions on Autonomous Mental Development},
  volume={7},
  number={3},
  pages={162--175},
  year={2015},
  publisher={IEEE}
}

@inproceedings{zafeiriou2017aff,
  title={Aff-wild: valence and arousal'In-the-Wild'challenge},
  author={Zafeiriou, Stefanos and Kollias, Dimitrios and Nicolaou, Mihalis A and Papaioannou, Athanasios and Zhao, Guoying and Kotsia, Irene},
  booktitle=CVPRW,
  pages={34--41},
  year={2017}
}

@article{katsigiannis2017dreamer,
  title={DREAMER: A database for emotion recognition through EEG and ECG signals from wireless low-cost off-the-shelf devices},
  author={Katsigiannis, Stamos and Ramzan, Naeem},
  journal={IEEE journal of Biomedical and Health Informatics},
  volume={22},
  number={1},
  pages={98--107},
  year={2017},
  publisher={IEEE}
}

@article{saganowski2022emognition,
  title={Emognition dataset: emotion recognition with self-reports, facial expressions, and physiology using wearables},
  author={Saganowski, Stanis{\l}aw and Komoszy{\'n}ska, Joanna and Behnke, Maciej and Perz, Bartosz and Kunc, Dominika and Klich, Bart{\l}omiej and Kaczmarek, {\L}ukasz D and Kazienko, Przemys{\l}aw},
  journal={Scientific Data},
  volume={9},
  number={1},
  pages={158},
  year={2022},
  publisher={Nature Publishing Group UK London}
}

@article{yang2024multimodal,
  title={A multimodal dataset for mixed emotion recognition},
  author={Yang, Pei and Liu, Niqi and Liu, Xinge and Shu, Yezhi and Ji, Wenqi and Ren, Ziqi and Sheng, Jenny and Yu, Minjing and Yi, Ran and Zhang, Dan and others},
  journal={Scientific Data},
  volume={11},
  number={1},
  pages={847},
  year={2024},
  publisher={Nature Publishing Group UK London}
}

@inproceedings{arnab2021vivit,
  title={Vivit: A video vision transformer},
  author={Arnab, Anurag and Dehghani, Mostafa and Heigold, Georg and Sun, Chen and Lu{\v{c}}i{\'c}, Mario and Schmid, Cordelia},
  booktitle=ICCV,
  pages={6836--6846},
  year={2021}
}

@inproceedings{bertasius2021space,
  title={Is space-time attention all you need for video understanding?},
  author={Bertasius, Gedas and Wang, Heng and Torresani, Lorenzo},
  booktitle=ICML,
  year={2021}
}

@article{li2020deep,
  title={Deep facial expression recognition: A survey},
  author={Li, Shan and Deng, Weihong},
  journal={IEEE transactions on affective computing},
  volume={13},
  number={3},
  pages={1195--1215},
  year={2020},
  publisher={IEEE}
}

@inproceedings{liu2022video,
  title={Video swin transformer},
  author={Liu, Ze and Ning, Jia and Cao, Yue and Wei, Yixuan and Zhang, Zheng and Lin, Stephen and Hu, Han},
  booktitle=CVPR,
  pages={3192--3201},
  year={2022},
  organization={IEEE}
}

@article{klem1999ten,
  title={The ten-twenty electrode system of the international federation. The international federation of clinical neurophysiology},
  author={Klem, George H},
  journal={Electroencephalogr. Clin. Neurophysiol. Suppl.},
  volume={52},
  pages={3--6},
  year={1999}
}

@article{tong2022videomae,
  title={Videomae: Masked autoencoders are data-efficient learners for self-supervised video pre-training},
  author={Tong, Zhan and Song, Yibing and Wang, Jue and Wang, Limin},
  journal=NIPS,
  volume={35},
  pages={10078--10093},
  year={2022}
}

@inproceedings{bandara2023adamae,
  title={Adamae: Adaptive masking for efficient spatiotemporal learning with masked autoencoders},
  author={Bandara, Wele Gedara Chaminda and Patel, Naman and Gholami, Ali and Nikkhah, Mehdi and Agrawal, Motilal and Patel, Vishal M},
  booktitle=CVPR,
  pages={14507--14517},
  year={2023}
}

@inproceedings{zhang2023crossformer,
  title={Crossformer: Transformer utilizing cross-dimension dependency for multivariate time series forecasting},
  author={Zhang, Yunhao and Yan, Junchi},
  booktitle=ICLR,
  year={2023}
}

@article{wang2024medformer,
  title={Medformer: A multi-granularity patching transformer for medical time-series classification},
  author={Wang, Yihe and Huang, Nan and Li, Taida and Yan, Yujun and Zhang, Xiang},
  journal=NIPS,
  volume={37},
  pages={36314--36341},
  year={2024}
}

@article{sun2024hicmae,
  title={Hicmae: Hierarchical contrastive masked autoencoder for self-supervised audio-visual emotion recognition},
  author={Sun, Licai and Lian, Zheng and Liu, Bin and Tao, Jianhua},
  journal={Information Fusion},
  volume={108},
  pages={102382},
  year={2024},
  publisher={Elsevier}
}

@article{gao2022eeg,
  title={EEG-GCN: Spatio-temporal and self-adaptive graph convolutional networks for single and multi-view EEG-based emotion recognition},
  author={Gao, Yue and Fu, Xiangling and Ouyang, Tianxiong and Wang, Yi},
  journal=SPL,
  volume={29},
  pages={1574--1578},
  year={2022},
  publisher={IEEE}
}

@article{du2022multi,
  title={A multi-dimensional graph convolution network for EEG emotion recognition},
  author={Du, Guanglong and Su, Jinshao and Zhang, Linlin and Su, Kang and Wang, Xueqian and Teng, Shaohua and Liu, Peter Xiaoping},
  journal={IEEE Transactions on Instrumentation and Measurement},
  volume={71},
  pages={1--11},
  year={2022},
  publisher={IEEE}
}

@article{pan2023msfr,
  title={MSFR-GCN: A multi-scale feature reconstruction graph convolutional network for EEG emotion and cognition recognition},
  author={Pan, Deng and Zheng, Haohao and Xu, Feifan and Ouyang, Yu and Jia, Zhe and Wang, Chu and Zeng, Hong},
  journal={IEEE Transactions on Neural Systems and Rehabilitation Engineering},
  volume={31},
  pages={3245--3254},
  year={2023},
  publisher={IEEE}
}

@article{ye2022hierarchical,
  title={Hierarchical dynamic graph convolutional network with interpretability for EEG-based emotion recognition},
  author={Ye, Mengqing and Chen, CL Philip and Zhang, Tong},
  journal={IEEE Transactions on Neural Networks and Learning Systems},
  year={2022},
  publisher={IEEE}
}

@article{jin2024pgcn,
  title={PGCN: Pyramidal graph convolutional network for EEG emotion recognition},
  author={Jin, Ming and Du, Changde and He, Huiguang and Cai, Ting and Li, Jinpeng},
  journal={IEEE Transactions on Multimedia},
  volume={26},
  pages={9070--9082},
  year={2024},
  publisher={IEEE}
}

@inproceedings{kipf2016semi,
  title={Semi-supervised classification with graph convolutional networks},
  author={Kipf, Thomas N and Welling, Max}, 
  booktitle = ICLR,
  year={2017}
}

@inproceedings{cui2019class,
  title={Class-balanced loss based on effective number of samples},
  author={Cui, Yin and Jia, Menglin and Lin, Tsung-Yi and Song, Yang and Belongie, Serge},
  booktitle=CVPR,
  pages={9268--9277},
  year={2019}
}

@article{el2026reve,
  title={REVE: A foundation model for EEG-adapting to any setup with large-scale pretraining on 25,000 subjects},
  author={El Ouahidi, Yassine and Lys, Jonathan and Th{\"o}lke, Philipp and Farrugia, Nicolas and Pasdeloup, Bastien and Gripon, Vincent and Jerbi, Karim and Lioi, Giulia},
  journal={Advances in Neural Information Processing Systems},
  volume={38},
  pages={22541--22577},
  year={2025}
}

@inproceedings{wang2025cbramod,
  title={Cbramod: A criss-cross brain foundation model for eeg decoding},
  author={Wang, Jiquan and Zhao, Sha and Luo, Zhiling and Zhou, Yangxuan and Jiang, Haiteng and Li, Shijian and Li, Tao and Pan, Gang},
  booktitle={International conference on learning representations},
  volume={2025},
  pages={75310--75346},
  year={2025}
}

@inproceedings{jiang2024labram,
  title={Large brain model for learning generic representations with tremendous EEG data in BCI},
  author={Jiang, Wei-Bang and Zhao, Liming and Lu, Bao-Liang},
  booktitle={International Conference on Learning Representations},
  volume={2024},
  pages={16405--16426},
  year={2024}
}

@article{yang2023biot,
  title={Biot: Biosignal transformer for cross-data learning in the wild},
  author={Yang, Chaoqi and Westover, M and Sun, Jimeng},
  journal={Advances in Neural Information Processing Systems},
  volume={36},
  pages={78240--78260},
  year={2023}
}

@article{song2021sttransformer,
  title={Transformer-based spatial-temporal feature learning for EEG decoding},
  author={Song, Yonghao and Jia, Xueyu and Yang, Lie and Xie, Longhan},
  journal={arXiv preprint arXiv:2106.11170},
  year={2021}
}

@article{wang2026milmer,
  title={Milmer: a framework for multiple instance learning based multimodal emotion recognition},
  author={Wang, Zaitian and He, Jian and Liang, Yu and Hu, Xiyuan and Peng, Tianhao and Wang, Kaixin and Wang, Jiakai and Zhang, Chenlong and Zhang, Weili and Niu, Shuang and others},
  journal={Neurocomputing},
  pages={134368},
  year={2026},
  publisher={Elsevier}
}

@inproceedings{piergiovanni2019temporal,
  title={Temporal gaussian mixture layer for videos},
  author={Piergiovanni, AJ and Ryoo, Michael},
  booktitle={International Conference on Machine learning},
  pages={5152--5161},
  year={2019},
  organization={PMLR}
}

@inproceedings{wang2018videos,
  title={Videos as space-time region graphs},
  author={Wang, Xiaolong and Gupta, Abhinav},
  booktitle={Proceedings of the European conference on computer vision (ECCV)},
  pages={399--417},
  year={2018}
}

@inproceedings{yang2020spatial,
  title={Spatial-temporal graph convolutional network for video-based person re-identification},
  author={Yang, Jinrui and Zheng, Wei-Shi and Yang, Qize and Chen, Ying-Cong and Tian, Qi},
  booktitle={Proceedings of the IEEE/CVF conference on computer vision and pattern recognition},
  pages={3289--3299},
  year={2020}
}

@inproceedings{yan2018spatial,
  title={Spatial temporal graph convolutional networks for skeleton-based action recognition},
  author={Yan, Sijie and Xiong, Yuanjun and Lin, Dahua},
  booktitle={Proceedings of the AAAI conference on artificial intelligence},
  volume={32},
  number={1},
  year={2018}
}

@inproceedings{nie2023patchtst,
  title={A Time Series is Worth 64 Words: Long-term Forecasting with Transformers},
  author={Nie, Yuqi and Nguyen, Nam H and Sinthong, Phanwadee and Kalagnanam, Jayant},
  booktitle={International Conference on Learning Representations},
  year={2023}
}

@inproceedings{zhou2021informer,
  title={Informer: Beyond efficient transformer for long sequence time-series forecasting},
  author={Zhou, Haoyi and Zhang, Shanghang and Peng, Jieqi and Zhang, Shuai and Li, Jianxin and Xiong, Hui and Zhang, Wancai},
  booktitle={Proceedings of the AAAI Conference on Artificial Intelligence},
  volume={35},
  number={12},
  pages={11106--11115},
  year={2021}
}

@inproceedings{zhou2022fedformer,
  title={FEDformer: Frequency enhanced decomposed transformer for long-term series forecasting},
  author={Zhou, Tian and Ma, Ziqing and Wen, Qingsong and Wang, Xue and Sun, Liang and Jin, Rong},
  booktitle={Proceedings of the International Conference on Machine Learning},
  pages={27268--27286},
  year={2022}
}

@article{siddhad2024eegtransformer,
  title={Efficacy of transformer networks for classification of raw EEG data},
  author={Siddhad, Gourav and Gupta, Anmol and Dogra, Debi Prosad and Roy, Partha Pratim},
  journal={Biomedical Signal Processing and Control},
  volume={87},
  pages={105488},
  year={2024},
  publisher={Elsevier}
}

@article{schirrmeister2017deep,
  title={Deep learning with convolutional neural networks for EEG decoding and visualization},
  author={Schirrmeister, Robin Tibor and Springenberg, Jost Tobias and Fiederer, Lukas Dominique Josef and Glasstetter, Martin and Eggensperger, Katharina and Tangermann, Michael and Hutter, Frank and Burgard, Wolfram and Ball, Tonio},
  journal={Human Brain Mapping},
  volume={38},
  number={11},
  pages={5391--5420},
  year={2017},
  publisher={Wiley}
}

@inproceedings{dosovitskiy2021vit,
  title={An image is worth 16x16 words: Transformers for image recognition at scale},
  author={Dosovitskiy, Alexey and Beyer, Lucas and Kolesnikov, Alexander and Weissenborn, Dirk and Zhai, Xiaohua and Unterthiner, Thomas and Dehghani, Mostafa and Minderer, Matthias and Heigold, Georg and Gelly, Sylvain and Uszkoreit, Jakob and Houlsby, Neil},
  booktitle={International Conference on Learning Representations},
  year={2021}
}
\bibliographystyle{iclr2027_conference}

\appendix
\section{Appendix}

\setcounter{figure}{0}
\setcounter{table}{0}
\renewcommand{\thefigure}{A\arabic{figure}}
\renewcommand{\thetable}{A\arabic{table}}

In this Appendix, we provide additional analyses, extended comparisons, and implementation details that could not be included in the main manuscript.
Specifically, we first evaluate the robustness of EVER under test-time EEG corruption and report the computational cost of all benchmarked models.
We further examine the statistical reliability of the benchmark through bootstrap confidence intervals, and analyze the complementarity between the video and EEG branches at the sample level.
Additional results include a comparison with six transformer architectures for multivariate time series adapted to EEG classification, together with the detailed experimental setup and the annotation distributions of the Emognition~\citep{saganowski2022emognition} and MDMER~\citep{yang2024multimodal} datasets.

\revb{Furthermore, Appendix~\ref{app:Evaluation Reliability} provides additional analyses of the evaluation, including the class distribution and the majority-class predictor (\Tabref{Tab:label_counts}), repeated runs over three seeds (\Tabref{Tab:multi_seed}), the paired comparison (\Tabref{Tab:paired_bootstrap}), the logit-averaging ensemble (\Tabref{Tab:ensemble}), the fusion comparison on both datasets and both backbone pairs (\Tabref{Tab:fusion_more} and \Figref{fig:cm_fusion}), the confusion matrices of the main models (\Figref{fig:cm_main}), the component ablation on Emognition (\Tabref{Tab:ablation_components_emog}), the ablation of the VCG prior on MDMER (\Tabref{Tab:ablation_vcg_mdmer}), and the within-region attention weights (\Tabref{Tab:alpha_distribution}).}

The source code will be made publicly available.

\subsection{Robustness to EEG Corruption}
\label{app:Robustness to EEG Corruption}

To evaluate robustness under unreliable EEG signals, we conduct a test-time corruption experiment on the Emognition dataset~\citep{saganowski2022emognition} without retraining the models.
For each EEG input $\mathbf{x}$, additive Gaussian noise is applied as follows:
\begin{equation}
\mathbf{x}'=\mathbf{x}+\boldsymbol{\varepsilon},
\qquad
\boldsymbol{\varepsilon}\sim
\mathcal{N}\left(\mathbf{0},(\lambda\sigma_{\mathbf{x}})^2\right),
\end{equation}
where $\sigma_{\mathbf{x}}$ denotes the standard deviation of the input EEG tensor and $\lambda\in\{0,0.1,0.2,0.3,0.5\}$ controls the corruption level.
The same corrupted EEG signal is provided to both the global CBraMod~\citep{wang2025cbramod} pathway and the STFT-based regional pathway, while the video input remains unchanged.
We compare the EEG-only CBraMod model with the AdaMAE-~\citep{bandara2023adamae} and VideoMAE-based~\citep{tong2022videomae} versions of EVER using Accuracy, UAR, and W-F1 averaged over valence and arousal.

As shown in \Tabref{Tab:eeg_noise_robustness}, CBraMod decreases from 0.353 to 0.294 in Accuracy as $\lambda$ increases from 0 to 0.5, corresponding to a degradation of 0.059 or 16.7\% of its clean performance.
In contrast, both versions of EVER degrade less, by \revb{0.030} for AdaMAE and \revb{0.036} for VideoMAE, and the VideoMAE-based fusion also maintains the highest UAR and W-F1 across all corruption levels.
The margin over the EEG-only model therefore widens with the corruption level, from \revb{0.018} to 0.047 in Accuracy for AdaMAE and from \revb{0.018} to 0.041 for VideoMAE.
Under the strongest corruption, both variants remain above the best video-only baseline (\ie, AdaMAE) of \revb{0.312}, showing that the fused prediction collapses to neither the degraded EEG branch nor the video branch alone.
Since the fusion gate is shared across samples, this robustness is attributed to multi-modal redundancy and to the inter-modal graph, in which the intact clip nodes continue to propagate information to the corrupted EEG nodes, rather than to sample-adaptive modality reweighting.

\begin{table*}[t!]
\caption{
\textbf{Robustness to test-time EEG corruption on the Emognition dataset.}
Each entry reports Acc/UAR/W-F1 averaged over valence and arousal.
}
\centering
\footnotesize
\resizebox{0.98\linewidth}{!}{
\begin{tabular}{c|c|ccccc|c}
\Xhline{3\arrayrulewidth}
\multirow{2}{*}{Method}
& \multirow{2}{*}{Backbone}
& \multicolumn{5}{c|}{\textbf{Noise level $\lambda$}}
& \multirow{2}{*}{$\Delta$ Acc} \\ \cline{3-7}
&
& 0.0 & 0.1 & 0.2 & 0.3 & 0.5
& \\ \hline\hline

EEG-only
& CBraMod
& 0.353/0.287/0.292
& 0.329/0.261/0.271
& 0.318/0.253/0.261
& 0.306/0.245/0.253
& 0.294/0.241/0.244
& $-0.059$ \\ \hline

Video + EEG
& AdaMAE
& \revb{0.371/0.306/0.298}
& 0.365/0.297/0.292
& 0.365/0.297/0.292
& 0.347/0.281/0.264
& 0.341/0.272/0.254
& \revb{$\mathbf{-0.030}$} \\

(Ours) & VideoMAE
& \revb{0.371}/\textbf{0.353}/\revb{\textbf{0.353}}
& \revb{0.376}/\textbf{0.350}/\textbf{0.348}
& 0.359/\textbf{0.347}/\textbf{0.335}
& 0.359/\textbf{0.347}/\textbf{0.334}
& 0.335/\textbf{0.315}/\textbf{0.304}
& \revb{$-0.036$} \\

\Xhline{3\arrayrulewidth}
\multicolumn{8}{r}{A smaller absolute $\Delta$ Acc indicates better robustness.}
\end{tabular}}
\label{Tab:eeg_noise_robustness}
\end{table*}

\subsection{Model Size and Accuracy}
\label{app:Model Size and Accuracy}

\begin{table*}[t!]
\centering
\caption{
\textbf{Model size and average performance on the Emognition and MDMER datasets.}
All performance values are averaged over valence and arousal.
}
\footnotesize
\resizebox{0.98\linewidth}{!}{
\begin{tabular}{c|c|c|ccc|ccc}
\Xhline{3\arrayrulewidth}
\multirow{2}{*}{Modality}
& \multirow{2}{*}{Model}
& \multirow{2}{*}{Params}
& \multicolumn{3}{c|}{\textbf{Emognition}}
& \multicolumn{3}{c}{\textbf{MDMER}} \\ \cline{4-9}
& & & Acc & UAR & W-F1 & Acc & UAR & W-F1 \\ \hline\hline

\multirow{5}{*}{Video}
& ViViT~\citep{arnab2021vivit}
& 88.7M
& \revb{0.241} & \revb{0.198} & \revb{0.214}
& 0.295 & \revb{0.223} & \revb{\underline{0.275}} \\

& Swin~\revb{\citep{liu2022video}}
& 87.6M
& \revb{0.259} & \revb{0.211} & \revb{0.224}
& 0.320 & 0.223 & \revb{0.272} \\

& TSF~\citep{bertasius2021space}
& 121.3M
& \revb{0.282} & \revb{0.227} & \revb{0.215}
& \revb{0.307} & \revb{0.208} & 0.224 \\

& VideoMAE~\citep{tong2022videomae}
& 86.4M
& \revb{0.300} & \revb{0.265} & \revb{0.284}
& 0.334 & \revb{0.219} & \revb{0.221} \\

& AdaMAE~\citep{bandara2023adamae}
& 87.6M
& \revb{0.312} & \revb{0.255} & \revb{0.231}
& 0.332 & 0.225 & 0.260 \\ \hline

\multirow{5}{*}{EEG}
& ST-Transformer~\citep{song2021sttransformer}
& 2.6M
& 0.318 & 0.267 & 0.259
& 0.251 & 0.185 & 0.201 \\

& BIOT~\citep{yang2023biot}
& 3.2M
& 0.324 & 0.285 & \revb{\underline{0.306}}
& 0.252 & 0.212 & 0.251 \\

& LaBraM~\citep{jiang2024labram}
& 5.8M
& 0.276 & 0.237 & 0.139
& 0.299 & 0.195 & 0.160 \\

& CBraMod~\citep{wang2025cbramod}
& 21.3M
& \revb{\underline{0.353}} & 0.287 & 0.292
& 0.302 & 0.208 & 0.219 \\

& REVE~\citep{el2026reve}
& 111.9M
& 0.276 & 0.238 & 0.240
& 0.283 & 0.205 & 0.234 \\ \hline

\multirow{4}{*}{Video+EEG}
& EAV~\citep{lee2024eav}
& 85.8M
& \revb{0.271} & 0.260 & \revb{0.203}
& \revb{0.334} & \revb{0.227} & \revb{0.231} \\

& Milmer~\citep{wang2026milmer}
& 1457.6M
& \revb{0.300} & \revb{0.256} & \revb{0.221}
& \revb{0.318} & \revb{0.213} & \revb{0.224} \\ \cline{2-9}

& \textbf{Ours(VideoMAE+CBraMod)}
& 135.8M
& \revb{\textbf{0.371}}
& \textbf{0.353}
& \revb{\textbf{0.353}}
& \revb{\textbf{0.354}}
& \revb{\textbf{0.237}}
& \revb{\textbf{0.280}} \\

& \textbf{Ours(AdaMAE+CBraMod)}
& 136.9M
& \revb{\textbf{0.371}}
& \revb{\underline{0.306}}
& \revb{0.298}
& \revb{\underline{0.341}}
& \revb{\underline{0.231}}
& \revb{0.274} \\

\Xhline{3\arrayrulewidth}
\multicolumn{9}{r}{
\textbf{Bold}: The best, \underline{Underline}: The second-best
}
\end{tabular}}
\label{Tab:model_size_performance}
\end{table*}

As shown in \Tabref{Tab:model_size_performance}, EVER provides a favorable trade-off between model capacity and recognition performance.
The two EVER variants contain 135.8M and 136.9M parameters, only about 9.3\% of Milmer's 1457.6M, and both outperform Milmer and EAV across all metrics on both datasets.
Since the paired backbones already account for 107.7M and 108.9M, the proposed modules add approximately 28M parameters, suggesting that parameter count alone does not explain the improvement.
Most EEG models are compact, ranging from 2.6M to 21.3M, with REVE as the main exception at 111.9M, which makes CBraMod a parameter-efficient global EEG backbone for fusion.

The relative effectiveness of the two modalities varies across datasets.
On Emognition, the EEG-only CBraMod model reaches 0.353 in average Accuracy and exceeds the best video-only baseline at \revb{0.312}, whereas on MDMER the best video model achieves 0.334 against 0.302 for the best EEG model.
This difference suggests that the informativeness of each modality depends on dataset-specific factors such as recording quality, subject variability, stimulus design, and the strength of observable facial responses.
Relying on a single modality is therefore unlikely to provide consistently strong performance across acquisition conditions.
EVER addresses this limitation by jointly exploiting complementary physiological and visual cues, and Ours (VideoMAE+CBraMod) improves over its own video and EEG backbones by \revb{0.071 and 0.018} on Emognition and by \revb{0.020 and 0.052} on MDMER.
Thus, the gain is obtained against whichever modality is weaker on a given dataset, showing that the fusion compensates for dataset-dependent weaknesses rather than inheriting the behavior of a single branch.

\subsection{Statistical Reliability of the Benchmark}
\label{app:Bootstrap Confidence Intervals}

Because paired EEG-video datasets are small and our splits are subject-independent, aggregate scores carry non-negligible sampling variance.
To quantify test-set sampling uncertainty for each fixed trained model, we compute percentile bootstrap confidence intervals from $B=10{,}000$ resamples of the test predictions in \Tabref{Tab:bootstrap_ci}, where all models are evaluated on identical resamples.

The interval widths follow the test sizes of the two datasets.
On Emognition with \revb{85} evaluation samples, the Accuracy interval of Ours (VideoMAE+CBraMod) spans \revb{0.153}, whereas the corresponding interval on MDMER with \revb{475} samples spans \revb{0.064}.
The Emognition intervals of the strongest models therefore overlap, and we do not claim that individual differences on this dataset are statistically separated.
On MDMER, the Accuracy interval of Ours (VideoMAE+CBraMod) is disjoint from those of ST-Transformer, BIOT, and REVE, and its W-F1 interval is disjoint from those of ST-Transformer, LaBraM, and CBraMod.
The intervals of the strongest video baselines remain partially overlapping, so the ranking among the top models should be interpreted as consistent rather than significant.
We regard this consistency as the more informative evidence, since EVER attains the \revb{best estimate in all six metric and dataset combinations}, whereas no baseline achieves this consistency.

\begin{table*}[t!]
\centering
\caption{
\textbf{Bootstrap 95\% confidence intervals on the Emognition and MDMER datasets.}
Intervals are percentile bootstrap CIs obtained from $B=10{,}000$ resamples,
averaged over valence and arousal.
}
\footnotesize
\resizebox{\linewidth}{!}{
\begin{tabular}{c|c|ccc|ccc}
\Xhline{3\arrayrulewidth}
\multirow{2}{*}{Modality}
& \multirow{2}{*}{Model}
& \multicolumn{3}{c|}{\textbf{Emognition}}
& \multicolumn{3}{c}{\textbf{MDMER}} \\ \cline{3-8}
& & Acc & UAR & W-F1 & Acc & UAR & W-F1 \\ \hline\hline
\multirow{5}{*}{Video}
& ViViT
& \revb{0.241 [.176,.312]} & \revb{0.198 [.144,.257]} & \revb{0.214 [.150,.285]}
& 0.295 [.266,.324] & \revb{0.223 [.198,.254]} & \revb{\underline{0.275} [.246,.306]} \\
& Swin
& \revb{0.259 [.194,.324]} & \revb{0.211 [.162,.262]} & \revb{0.224 [.165,.288]}
& \revb{0.320 [.289,.351]} & 0.223 [.204,.243] & \revb{0.272 [.242,.302]} \\
& TSF
& \revb{0.282 [.218,.347]} & \revb{0.227 [.182,.273]} & \revb{0.215 [.154,.282]}
& \revb{0.307 [.277,.337]} & \revb{0.208 [.190,.226]} & \revb{0.224 [.194,.255]} \\
& VideoMAE
& \revb{0.300 [.224,.376]} & \revb{0.265 [.200,.331]} & \revb{0.284 [.208,.363]}
& 0.334 [.303,.365] & \revb{0.219 [.207,.232]} & \revb{0.221 [.192,.251]} \\
& AdaMAE
& \revb{0.312 [.241,.382]} & \revb{0.255 [.205,.302]} & \revb{0.231 [.168,.303]}
& \revb{0.332 [.301,.362]} & \revb{0.225 [.207,.243]} & \revb{0.260 [.231,.289]} \\ \hline
\multirow{5}{*}{EEG}
& ST-Transformer
& \revb{0.318 [.253,.382]} & \revb{0.267 [.214,.323]} & \revb{0.259 [.195,.328]}
& \revb{0.251 [.224,.277]} & 0.185 [.163,.209] & \revb{0.201 [.175,.227]} \\
& BIOT
& \revb{0.324 [.247,.400]} & \revb{0.285 [.217,.356]} & \revb{\underline{0.306} [.228,.387]}
& \revb{0.252 [.223,.281]} & \revb{0.212 [.182,.248]} & \revb{0.251 [.220,.282]} \\
& LaBraM
& \revb{0.276 [.200,.359]} & \revb{0.237 [.208,.264]} & \revb{0.139 [.085,.210]}
& \revb{0.299 [.268,.329]} & \revb{0.195 [.185,.204]} & 0.160 [.135,.187] \\
& CBraMod
& \revb{\underline{0.353} [.276,.429]} & \revb{0.287 [.229,.347]} & \revb{0.292 [.216,.369]}
& \revb{0.302 [.276,.327]} & \revb{0.208 [.193,.224]} & \revb{0.219 [.193,.245]} \\
& REVE
& \revb{0.276 [.206,.347]} & \revb{0.238 [.183,.297]} & \revb{0.240 [.172,.312]}
& \revb{0.283 [.255,.313]} & 0.205 [.182,.232] & \revb{0.234 [.206,.264]} \\ \hline
\multirow{4}{*}{Video+EEG}
& EAV
& \revb{0.271 [.194,.347]} & \revb{0.260 [.198,.322]} & \revb{0.203 [.135,.281]}
& \revb{0.334 [.302,.365]} & \revb{0.227 [.211,.242]} & \revb{0.231 [.199,.263]} \\
& Milmer
& \revb{0.300 [.229,.376]} & \revb{0.256 [.197,.316]} & \revb{0.221 [.154,.297]}
& \revb{0.318 [.287,.349]} & \revb{0.213 [.197,.229]} & \revb{0.224 [.194,.255]} \\ \cline{2-8}
& \textbf{Ours(VideoMAE+CBraMod)}
& \revb{\textbf{0.371} [.294,.447]} & \revb{\textbf{0.353} [.272,.435]} & \revb{\textbf{0.353} [.273,.433]}
& \revb{\textbf{0.354} [.321,.385]} & \revb{\textbf{0.237} [.220,.255]} & \revb{\textbf{0.280} [.248,.313]} \\
& \textbf{Ours(AdaMAE+CBraMod)}
& \revb{\textbf{0.371} [.294,.447]} & \revb{\underline{0.306} [.252,.360]} & \revb{0.298 [.223,.377]}
& \revb{\underline{0.341} [.311,.372]} & \revb{\underline{0.231} [.212,.250]} & \revb{0.274 [.245,.305]} \\
\Xhline{3\arrayrulewidth}
\multicolumn{8}{r}{
\textbf{Bold}: The best, \underline{Underline}: The second-best
}
\end{tabular}}
\label{Tab:bootstrap_ci}
\end{table*}

\subsection{Complementarity of EEG and Video}
\label{app:Complementarity}

\begin{figure*}[ht]
    \centering
    \renewcommand{\arraystretch}{0.2}
    \includegraphics[width=0.8\linewidth]{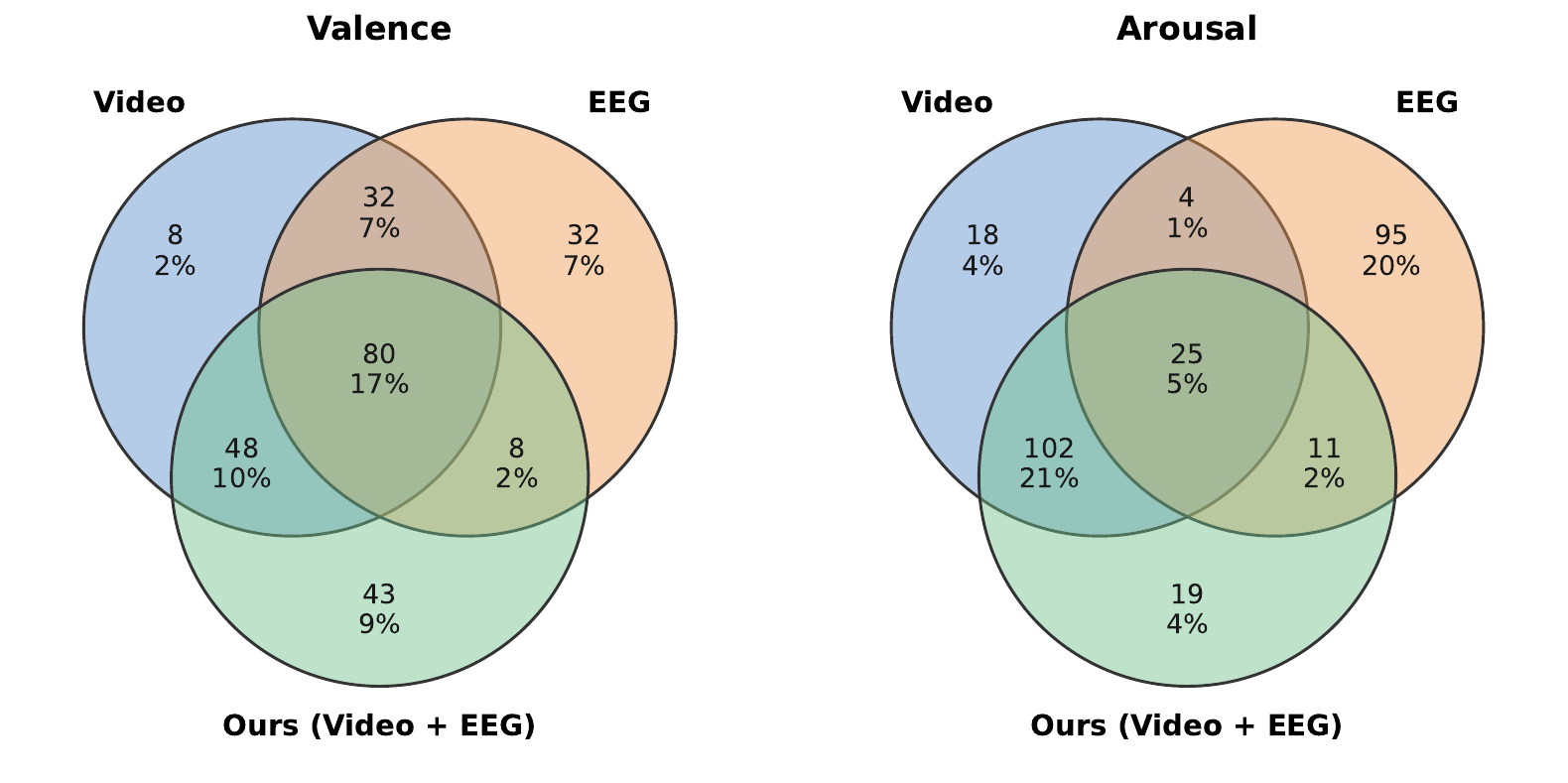}
    \caption{
    \textbf{Complementary correctness of the video, EEG, and fused predictions on the MDMER dataset.}
    Each region reports the number of test samples and its share of the evaluation set.
    \revb{The three models are VideoMAE, CBraMod, and EVER (VideoMAE+CBraMod) of \Tabref{Tab:emognition_mdmer_benchmark}, evaluated on the 475 test trials.}
    }
    \label{fig:venn_mdmer}
    \end{figure*}

{
To examine how the predictions of EVER relate to the predictions of its backbones, we compare the correctly classified sets of VideoMAE, CBraMod, and EVER (VideoMAE+CBraMod) on the 475 MDMER test trials in \Figref{fig:venn_mdmer}.
EVER improves over the better unimodal model from 35.4\% to 37.7\% on valence and from 31.4\% to 33.1\% on arousal.
The two unimodal models disagree on 20\% of the trials for valence and 48\% for arousal, which is related to the class on which each model concentrates its predictions.
We interpret the overlap statistics as a description of the predictions rather than as direct evidence of complementary information.

On valence, EVER assigns 175 trials to a class that VideoMAE and CBraMod predict for only 65 and 34 trials, and it classifies 43 trials correctly although both unimodal models fail.
On arousal, video is the stronger modality and EVER preserves 102 of the 120 trials solved only by VideoMAE, while the contribution of the EEG branch is limited.
Thus, EVER does not reproduce the prediction of a single backbone, and the quantitative benefit of fusion is reported in \Tabref{Tab:emognition_mdmer_benchmark}, \Tabref{Tab:heldout_mdmer}, and \Tabref{Tab:ensemble}.
}

\subsection{Video Clip Configuration}
\label{app:Video Clip Configuration}

\begin{table}[t!]
\centering
\caption{
\textbf{Effect of the clip sampling configuration on the Emognition dataset.}
Both settings use the AdaMAE backbone with a learning rate of $1\times10^{-4}$ and a weight decay of $0.05$.
}
\footnotesize
\resizebox{0.6\linewidth}{!}{
\begin{tabular}{c|cc|ccc}
\Xhline{3\arrayrulewidth}
\multirow{2}{*}{Backbone} & \multirow{2}{*}{Frame interval} & \multirow{2}{*}{Overlap}
& \multicolumn{3}{c}{\textbf{Emognition}} \\ \cline{4-6}
& & & Acc & UAR & W-F1 \\ \hline\hline
\multirow{2}{*}{AdaMAE}
& 2 & -- & \revb{0.276} & \revb{0.240} & \revb{0.209} \\
& 4 (ours) & 0.5 & \revb{\textbf{0.312}} & \revb{\textbf{0.255}} & \revb{\textbf{0.231}} \\
\Xhline{3\arrayrulewidth}
\end{tabular}}
\label{Tab:clip_config}
\end{table}

\begin{table*}[t!]
\centering
\caption{
\textbf{Comparison with transformer architectures for multivariate time series, adapted to EEG classification.}
We use the classification-adapted implementations from the Medformer benchmark~\citep{wang2024medformer}.
All values are averaged over valence and arousal, and all models are trained under the same subject-independent splits and training budget as the main benchmark.
}
\footnotesize
\resizebox{0.95\linewidth}{!}{
\begin{tabular}{c|c|ccc|ccc}
\Xhline{3\arrayrulewidth}
\multirow{2}{*}{Modality}
& \multirow{2}{*}{Model}
& \multicolumn{3}{c|}{\textbf{Emognition}}
& \multicolumn{3}{c}{\textbf{MDMER}} \\ \cline{3-8}
& & Acc & UAR & W-F1 & Acc & UAR & W-F1 \\ \hline\hline

\multirow{6}{*}{EEG}
& PatchTST~\citep{nie2023patchtst}
& 0.259 & 0.210 & 0.219
& 0.281 & 0.204 & \underline{0.266} \\

& Informer~\citep{zhou2021informer}
& \underline{0.318} & 0.241 & 0.208
& 0.300 & 0.210 & 0.215 \\

& FEDformer~\citep{zhou2022fedformer}
& 0.282 & \underline{0.270} & \underline{0.275}
& 0.264 & 0.195 & 0.244 \\

& Crossformer~\citep{zhang2023crossformer}
& 0.282 & 0.214 & 0.221
& \underline{0.322} & 0.205 & 0.184 \\

& Medformer~\citep{wang2024medformer}
& 0.312 & 0.223 & 0.206
& 0.318 & 0.212 & 0.230 \\

& EEG-Transformer~\citep{siddhad2024eegtransformer}
& 0.300 & 0.212 & 0.171
& 0.302 & \underline{0.217} & 0.200 \\ \hline

\multirow{2}{*}{Video+EEG}
& \textbf{Ours (VideoMAE+CBraMod)}
& \revb{\textbf{0.371}} & \textbf{0.353} & \revb{\textbf{0.353}}
& \revb{\textbf{0.354}} & \revb{\textbf{0.237}} & \revb{\textbf{0.280}} \\

& \textbf{Ours (AdaMAE+CBraMod)}
& \revb{\textbf{0.371}} & \revb{0.306} & \revb{0.298}
& \revb{0.341} & \revb{0.231} & \revb{0.274} \\

\Xhline{3\arrayrulewidth}
\multicolumn{8}{r}{
\textbf{Bold}: The best, \underline{Underline}: The best among the additional EEG models
}
\end{tabular}}
\label{Tab:additional_eeg}
\end{table*}

\begin{table*}[t!]
\centering
\caption{
\textbf{Model-specific learning rate and weight decay.}
All other training settings are shared across models.
}
\footnotesize
\resizebox{0.99\linewidth}{!}{
\begin{tabular}{c|c|cc}
\Xhline{3\arrayrulewidth}
Modality & Model & LR & Weight decay \\ \hline\hline
\multirow{2}{*}{Video}
& ViViT, Swin, TSF & $2\times10^{-5}$ & 0.02 \\
& VideoMAE, AdaMAE & $1\times10^{-4}$ & 0.05 \\ \hline
\multirow{2}{*}{EEG}
& ST-Transformer, BIOT, LaBraM, CBraMod, REVE & $1\times10^{-4}$ & 0.05 \\
& PatchTST, Informer, FEDformer, Crossformer, Medformer, EEG-Transformer & $1\times10^{-4}$ & 0.05 \\ \hline
Video+EEG
& \textbf{Ours (VideoMAE/AdaMAE + CBraMod)} & $1\times10^{-4}$ & 0.05 \\
\Xhline{3\arrayrulewidth}
\end{tabular}}
\label{Tab:implementation_hparams}
\end{table*}

All video models in the benchmark, including EVER, sample clips with a frame interval of four and an overlap ratio of $0.5$.
\Tabref{Tab:clip_config} compares this setting against non-overlapping clips with a frame interval of two on the AdaMAE backbone under an identical optimization setting.
The two configurations advance by the same hop of 64 frames and therefore yield the same number of clips and the same overall coverage of a recording, but each of our clips spans twice the temporal extent and shares half of its content with its neighbors.
This widens the temporal context available to a single clip and keeps adjacent clips continuous rather than disjoint, improving Accuracy from \revb{0.276 to 0.312} with consistent gains in UAR and W-F1 at no additional computational cost.
Thus, we adopt it for every video backbone, so that the reported gains of EVER are obtained against baselines evaluated under their stronger setting.

\subsection{Additional EEG Backbones}
\label{app:Additional EEG Backbones}

The EEG models in the main benchmark are either EEG foundation models or transformers designed for general EEG decoding.
To examine whether the advantage of EVER persists against a different family of EEG encoders, we additionally benchmark six transformer architectures for multivariate time series, namely PatchTST~\citep{nie2023patchtst}, Informer~\citep{zhou2021informer}, FEDformer~\citep{zhou2022fedformer}, Crossformer~\citep{zhang2023crossformer}, Medformer~\citep{wang2024medformer}, and EEG-Transformer~\citep{siddhad2024eegtransformer}.
The first four were originally proposed for long-term forecasting, whereas the latter two target medical and EEG signals directly.
We adopt the classification-adapted implementations released with the Medformer benchmark~\citep{wang2024medformer}, which reformulates these forecasting architectures for medical time-series classification on EEG and ECG recordings.
Since they encode multi-channel EEG through the same input interface as our benchmark, they can be trained under identical subject-independent splits, loss formulation, and training budget, and therefore serve as an independent set of EEG-only references obtained without any modification of our evaluation protocol.

As shown in \Tabref{Tab:additional_eeg}, these models perform within the range of the EEG backbones reported in the main paper.
On Emognition, their average Accuracy ranges from 0.259 to 0.318, and on MDMER from 0.264 to 0.322.
Crossformer reaches 0.322 on MDMER and thereby exceeds the strongest EEG model of the main benchmark at 0.302, which indicates that these architectures are competitive EEG encoders rather than weak references.
Nevertheless, all six remain below both versions of EVER on every metric and dataset.
The gap is largest for the class-balanced metrics on Emognition, where EVER improves UAR from 0.270 to 0.353 and W-F1 from 0.275 to \revb{0.353} relative to the best of these models.
Since the six models span frequency-domain, patch-based, and cross-dimensional attention designs, their consistent shortfall suggests that the limitation of the EEG-only setting is the absence of the complementary visual modality rather than the choice of EEG architecture.
Together with the corruption experiment in Sec.~\ref{app:Robustness to EEG Corruption}, this supports that the benefit of Ours (Video + EEG) is robust to both the quality of the EEG signal and the design of the EEG encoder.
All six models use the classification implementations of the Medformer codebase~\citep{wang2024medformer} and are trained with the same learning rate and weight decay as the EEG backbones of the main benchmark (\Tabref{Tab:implementation_hparams}), under an identical optimizer, schedule, loss, and training budget.

\subsection{Additional Implementation Details}
\label{app:Implementation Details}

\begin{figure*}[ht]
    \centering
    \renewcommand{\arraystretch}{0.2}
    \begin{tabular}{c@{\hskip 0.003\linewidth}c}
    \includegraphics[width=0.45\linewidth]{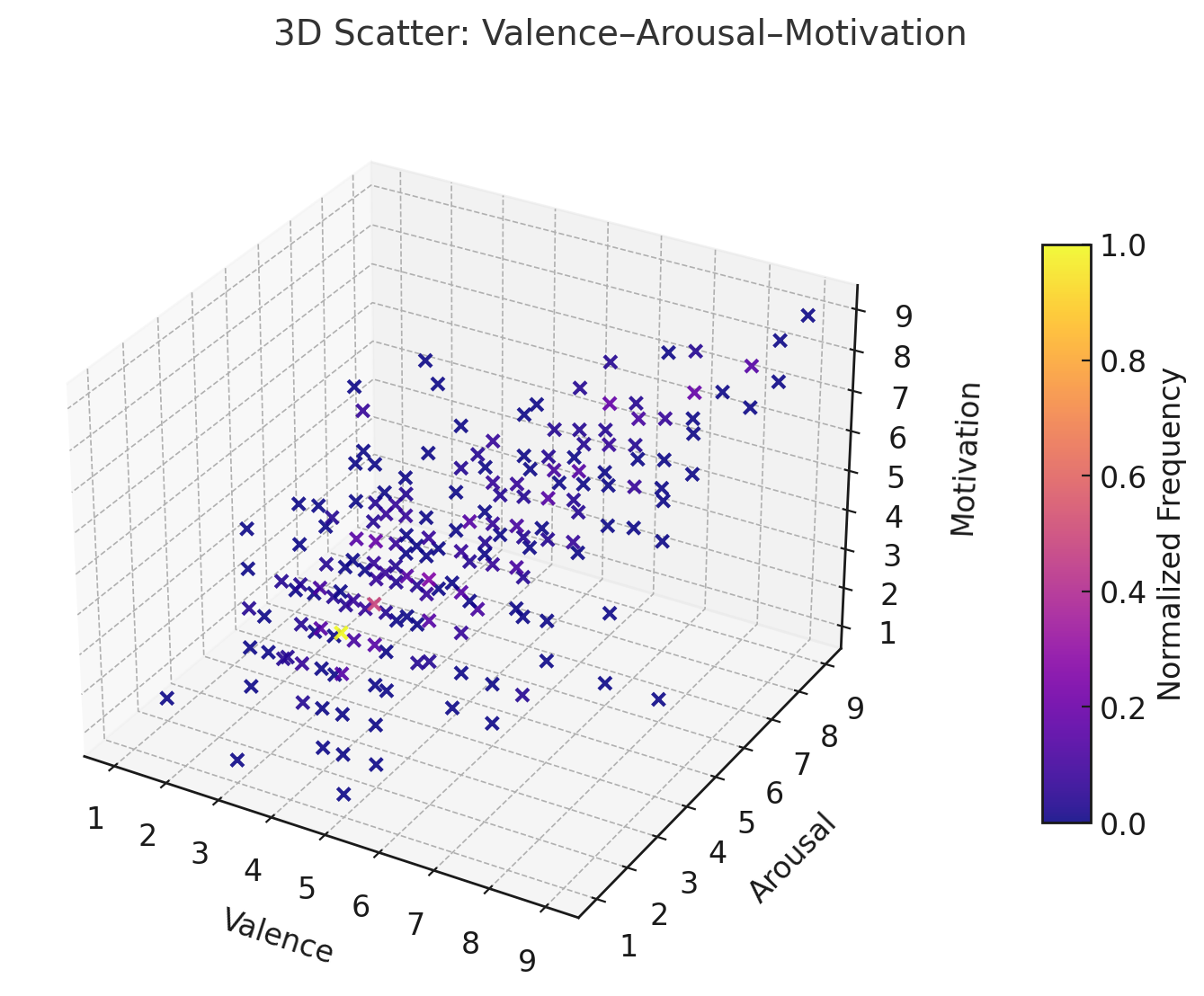} &
    \includegraphics[width=0.45\linewidth]{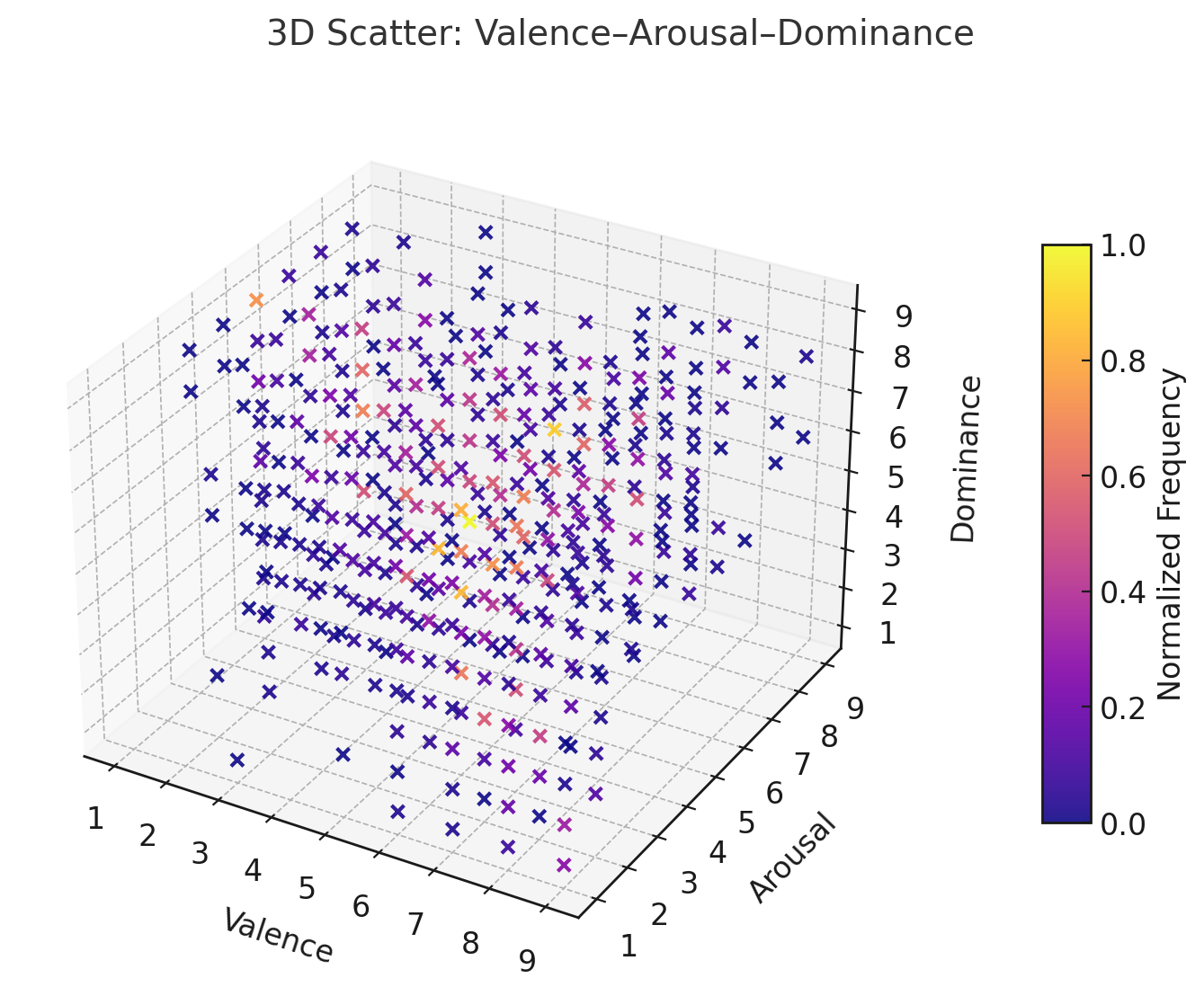} \\
    \footnotesize Emognition dataset & \footnotesize MDMER dataset
    \end{tabular}
    \caption{Distribution of self-assessment annotations (\ie, ground-truth scores) on the Emognition and MDMER datasets.}
    \label{fig:dataset_scatter}
\end{figure*}

\subsubsection{Experimental Setup}
\label{app:Experimental Setup}

Although we have already described the environment setup in the main manuscript, we provide additional details of the benchmark configuration.
We employ the AdamW optimizer with $\beta=(0.9,0.999)$ and $\epsilon=10^{-8}$.
The learning rate follows a linear warmup over the first 10\% of the total training steps, after which it is decayed by a factor of $0.75$ every ten epochs starting from the tenth epoch, implemented through a LambdaLR scheduler.
All networks are trained for 100 epochs with a batch size of 12, and training is terminated early when the \revb{average accuracy on the evaluation split} does not improve for 15 consecutive epochs.
All experiments are conducted on NVIDIA RTX A6000 GPUs with PyTorch 2.1.1 and CUDA 11.8.

For video, we adopt dense clip sampling with a frame interval of four and an overlap ratio of $0.5$ between consecutive clips, and the frame rate is normalized across recordings so that clip durations remain comparable between datasets.
Only the last block of each pretrained video backbone is unfrozen, while the remaining blocks are kept frozen to preserve large-scale pretrained representations under the limited size of paired EEG-video data.
In contrast to the video branch, all EEG backbones are fully fine-tuned, since EEG foundation models are pretrained on recording conditions that differ substantially from the affective stimuli considered here.

Learning rates and weight decay follow the configurations recommended for each backbone, as summarized in \Tabref{Tab:implementation_hparams}.
The supervised video transformers are trained with a smaller learning rate of $2\times10^{-5}$ and a weight decay of $0.02$, whereas the masked-autoencoder backbones, all EEG backbones, and EVER use a learning rate of $1\times10^{-4}$ and a weight decay of $0.05$.
EVER inherits the optimization setting of its paired backbones, so that the comparison against the unimodal models isolates the effect of the proposed fusion rather than differences in the optimization budget.

\paragraph{Multi-modal Baselines.}
Milmer~\citep{wang2026milmer} is a multi-modal architecture and is therefore evaluated as released.
EAV~\citep{lee2024eav}, in contrast, is a dataset paper that provides modality-specific baselines but no joint fusion model.
To obtain a faithful Video+EEG comparison, we pair its two published branches with a standard late-fusion head.
The EEG branch follows the released ShallowConvNet~\citep{schirrmeister2017deep} baseline: the EEG signal is downsampled to the 100\,Hz target rate of EAV, band-pass filtered between 0.5 and 45\,Hz, and segmented into five-second windows of 500 samples before the convolutional encoder.
The video branch follows the per-frame ViT-B/16~\citep{dosovitskiy2021vit} protocol of EAV, in which every frame is classified independently and the frame-level logits are averaged into a sequence-level prediction.
The two branches are combined by concatenating their logits and applying a single linear layer, which is the standard decision-level fusion used in the literature, and the model is trained under the subject-independent protocol as all other benchmark.

\subsubsection{Details of Dataset}
\label{app:Details of Dataset}

\begin{figure*}[ht]
    \centering
    \renewcommand{\arraystretch}{0.2}
    \begin{tabular}{c@{\hskip 0.003\linewidth}c@{\hskip 0.003\linewidth}c@{\hskip 0.003\linewidth}c@{\hskip 0.003\linewidth}c@{\hskip 0.003\linewidth}c@{\hskip 0.003\linewidth}c@{\hskip 0.003\linewidth}c}
    \includegraphics[width=0.12\linewidth]{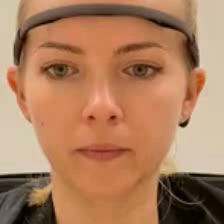} &
    \includegraphics[width=0.12\linewidth]{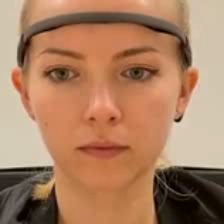} &
    \includegraphics[width=0.12\linewidth]{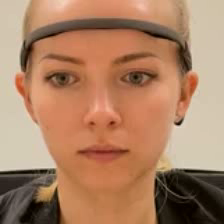} &
    \includegraphics[width=0.12\linewidth]{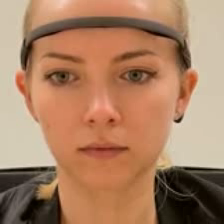} &
    \includegraphics[width=0.12\linewidth]{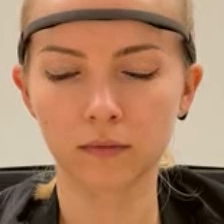} &
    \includegraphics[width=0.12\linewidth]{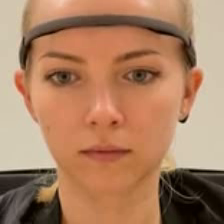} &
    \includegraphics[width=0.12\linewidth]{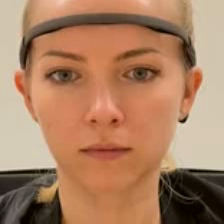} &
    \includegraphics[width=0.12\linewidth]{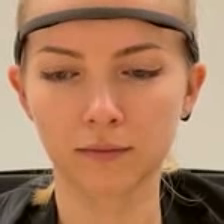}
    \\
    \footnotesize frame 1 & \footnotesize frame 2 & \footnotesize frame 3 & \footnotesize frame 4 & \footnotesize frame 5 & \footnotesize frame 6 & \footnotesize frame 7 & \footnotesize frame 8
    \\
    \includegraphics[width=0.12\linewidth]{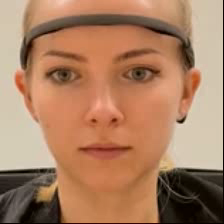} &
    \includegraphics[width=0.12\linewidth]{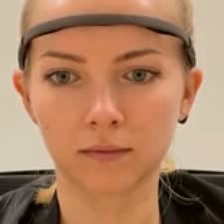} &
    \includegraphics[width=0.12\linewidth]{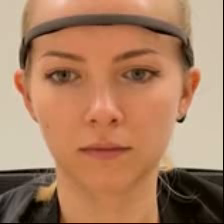} &
    \includegraphics[width=0.12\linewidth]{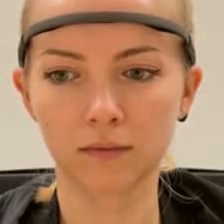} &
    \includegraphics[width=0.12\linewidth]{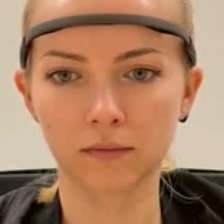} &
    \includegraphics[width=0.12\linewidth]{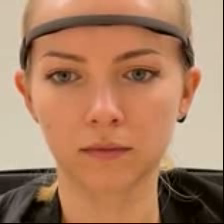} &
    \includegraphics[width=0.12\linewidth]{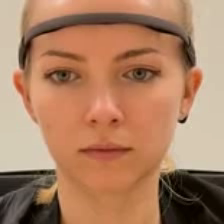} &
    \includegraphics[width=0.12\linewidth]{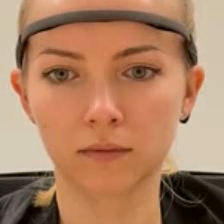}
    \\
    \footnotesize frame 9 & \footnotesize frame 10 & \footnotesize frame 11 & \footnotesize frame 12 & \footnotesize frame 13 & \footnotesize frame 14 & \footnotesize frame 15 & \footnotesize frame 16
    \\
    \includegraphics[width=0.12\linewidth]{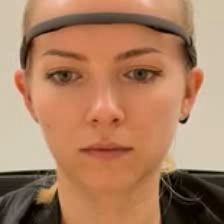} &
    \includegraphics[width=0.12\linewidth]{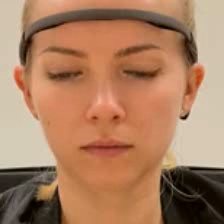} &
    \includegraphics[width=0.12\linewidth]{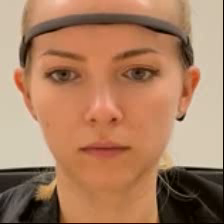} &
    \includegraphics[width=0.12\linewidth]{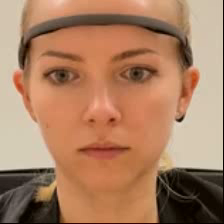} &
    \includegraphics[width=0.12\linewidth]{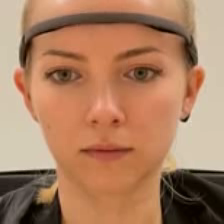} &
    \includegraphics[width=0.12\linewidth]{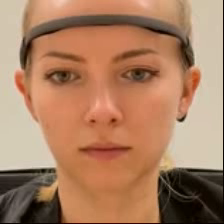} &
    \includegraphics[width=0.12\linewidth]{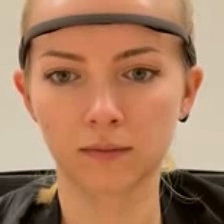} &
    \includegraphics[width=0.12\linewidth]{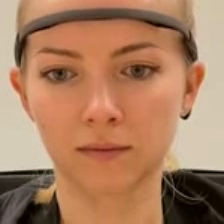}
    \\
    \footnotesize frame 17 & \footnotesize frame 18 & \footnotesize frame 19 & \footnotesize frame 20 & \footnotesize frame 21 & \footnotesize frame 22 & \footnotesize frame 23 & \footnotesize frame 24
    \\
    \includegraphics[width=0.12\linewidth]{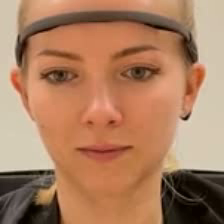} &
    \includegraphics[width=0.12\linewidth]{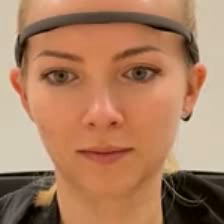} &
    \includegraphics[width=0.12\linewidth]{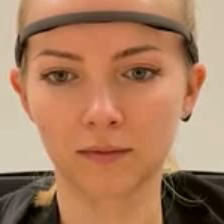} &
    \includegraphics[width=0.12\linewidth]{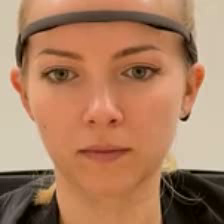} &
    \includegraphics[width=0.12\linewidth]{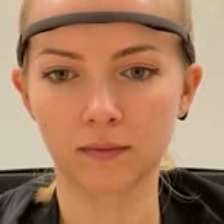} &
    \includegraphics[width=0.12\linewidth]{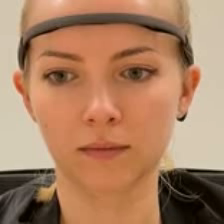} &
    \includegraphics[width=0.12\linewidth]{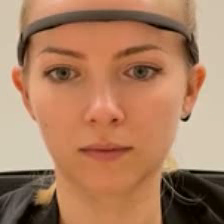} &
    \includegraphics[width=0.12\linewidth]{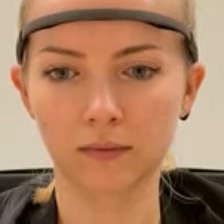}
    \\
    \footnotesize frame 25 & \footnotesize frame 26 & \footnotesize frame 27 & \footnotesize frame 28 & \footnotesize frame 29 & \footnotesize frame 30 & \footnotesize frame 31 & \footnotesize frame 32
    \end{tabular}
    \caption{Sample of video frames on the Emognition dataset with disgust stimuli.}
    \label{fig:emognition_frames}
\end{figure*}

\begin{figure*}[ht]
    \centering
    \renewcommand{\arraystretch}{0.2}
    \begin{tabular}{c@{\hskip 0.003\linewidth}c@{\hskip 0.003\linewidth}c@{\hskip 0.003\linewidth}c@{\hskip 0.003\linewidth}c@{\hskip 0.003\linewidth}c@{\hskip 0.003\linewidth}c@{\hskip 0.003\linewidth}c}
    \includegraphics[width=0.12\linewidth]{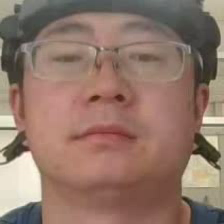} &
    \includegraphics[width=0.12\linewidth]{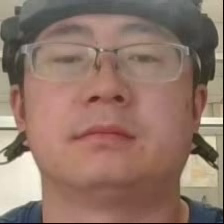} &
    \includegraphics[width=0.12\linewidth]{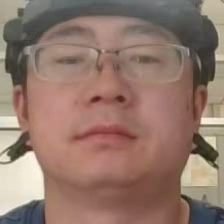} &
    \includegraphics[width=0.12\linewidth]{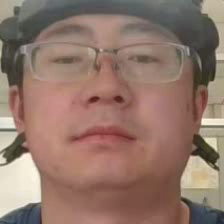} &
    \includegraphics[width=0.12\linewidth]{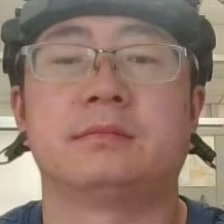} &
    \includegraphics[width=0.12\linewidth]{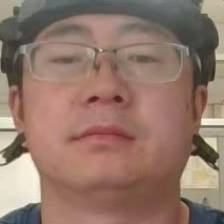} &
    \includegraphics[width=0.12\linewidth]{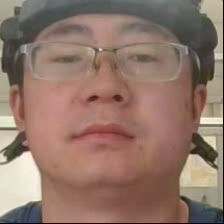} &
    \includegraphics[width=0.12\linewidth]{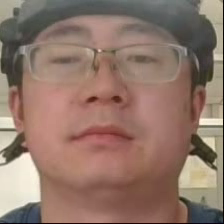}
    \\
    \footnotesize frame 1 & \footnotesize frame 2 & \footnotesize frame 3 & \footnotesize frame 4 & \footnotesize frame 5 & \footnotesize frame 6 & \footnotesize frame 7 & \footnotesize frame 8
    \\
    \includegraphics[width=0.12\linewidth]{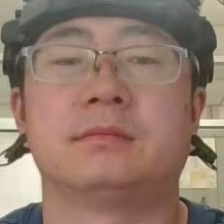} &
    \includegraphics[width=0.12\linewidth]{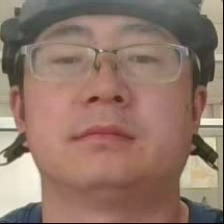} &
    \includegraphics[width=0.12\linewidth]{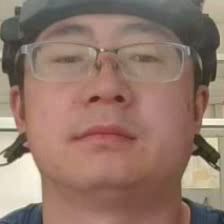} &
    \includegraphics[width=0.12\linewidth]{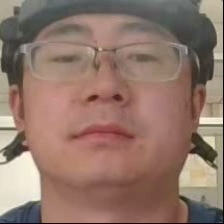} &
    \includegraphics[width=0.12\linewidth]{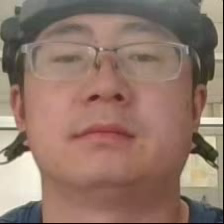} &
    \includegraphics[width=0.12\linewidth]{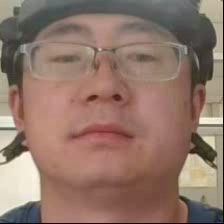} &
    \includegraphics[width=0.12\linewidth]{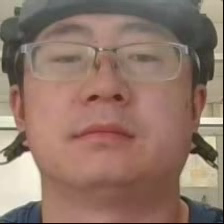} &
    \includegraphics[width=0.12\linewidth]{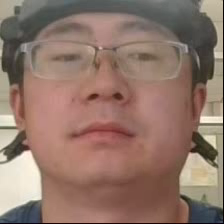}
    \\
    \footnotesize frame 9 & \footnotesize frame 10 & \footnotesize frame 11 & \footnotesize frame 12 & \footnotesize frame 13 & \footnotesize frame 14 & \footnotesize frame 15 & \footnotesize frame 16
    \\
    \includegraphics[width=0.12\linewidth]{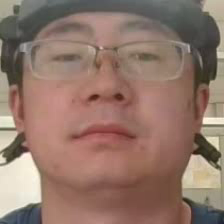} &
    \includegraphics[width=0.12\linewidth]{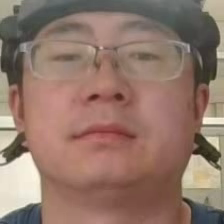} &
    \includegraphics[width=0.12\linewidth]{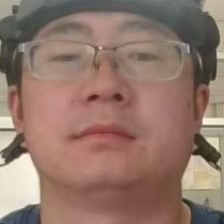} &
    \includegraphics[width=0.12\linewidth]{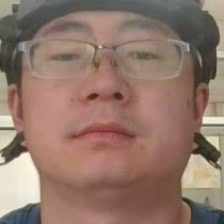} &
    \includegraphics[width=0.12\linewidth]{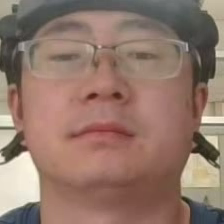} &
    \includegraphics[width=0.12\linewidth]{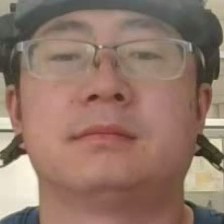} &
    \includegraphics[width=0.12\linewidth]{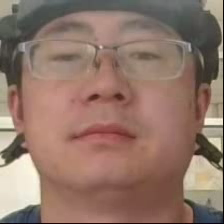} &
    \includegraphics[width=0.12\linewidth]{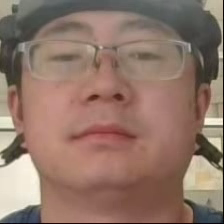}
    \\
    \footnotesize frame 17 & \footnotesize frame 18 & \footnotesize frame 19 & \footnotesize frame 20 & \footnotesize frame 21 & \footnotesize frame 22 & \footnotesize frame 23 & \footnotesize frame 24
    \\
    \includegraphics[width=0.12\linewidth]{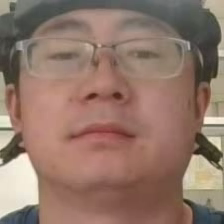} &
    \includegraphics[width=0.12\linewidth]{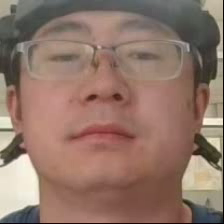} &
    \includegraphics[width=0.12\linewidth]{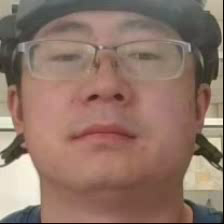} &
    \includegraphics[width=0.12\linewidth]{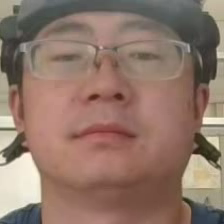} &
    \includegraphics[width=0.12\linewidth]{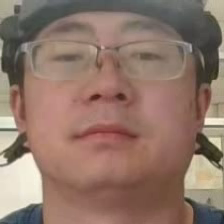} &
    \includegraphics[width=0.12\linewidth]{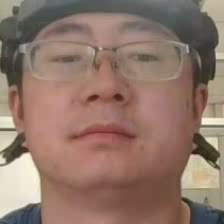} &
    \includegraphics[width=0.12\linewidth]{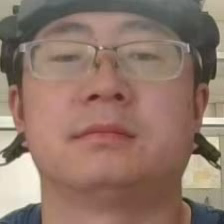} &
    \includegraphics[width=0.12\linewidth]{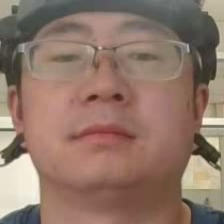}
    \\
    \footnotesize frame 25 & \footnotesize frame 26 & \footnotesize frame 27 & \footnotesize frame 28 & \footnotesize frame 29 & \footnotesize frame 30 & \footnotesize frame 31 & \footnotesize frame 32
    \end{tabular}
    \caption{Sample of video frames on the MDMER dataset.}
    \label{fig:mdmer_frames}
\end{figure*}

To further describe the details of the datasets, we visualize the 3D scatter plots of the self-assessment annotations for the Emognition~\citep{saganowski2022emognition} and MDMER~\citep{yang2024multimodal} datasets as shown in Fig. \ref{fig:dataset_scatter}.
Each axis in the visualization corresponds to the respective annotation dimensions. The Emognition dataset is originally annotated along the dimensions of valence–arousal–motivation, while MDMER is based on valence–arousal–dominance. 
We observe that Emognition annotations are unevenly distributed across the space, whereas MDMER covers a wider range but remains concentrated near the middle scores.
To address such imbalance and concentration issues during training, we adopt a combination of weighted cross-entropy loss and focal loss.
We focus our prediction on valence and arousal, which are the most widely adopted dimensions in emotion research (\eg, the circumplex model of affect) and serve as a common, interpretable basis for evaluating emotional states.
\Figref{fig:emognition_frames} and \Figref{fig:mdmer_frames} show representative 32-frame clips from the two datasets, illustrating that facial responses evolve gradually within a clip and motivating the clip-level temporal modeling of the VCG prior.

{
\subsection{Additional Analyses of Evaluation Reliability}
\label{app:Evaluation Reliability}

This section collects the analyses that complement the benchmark in the main paper.
Unless stated otherwise, all analyses are conducted under the evaluation protocol described in \secref{subsubsec:Environments}.

\paragraph{Class distribution and majority-class predictor.}
\Tabref{Tab:label_counts} reports the five-class label counts of both datasets.
A predictor that always outputs the most frequent class obtains an average accuracy of 0.300 on Emognition and 0.317 on MDMER, with a UAR of 0.200 in both cases.
These values provide the reference level for the accuracies in \Tabref{Tab:emognition_mdmer_benchmark}.

\begin{table}[h!]
\centering
\caption{\revb{\textbf{Five-class label counts (classes 0 to 4) and majority-class accuracy on the test set.}}}
\footnotesize
\resizebox{0.85\linewidth}{!}{
\begin{tabular}{c|c|c|c}
\Xhline{3\arrayrulewidth}
Dataset, target & Training set & Test set & Majority-class Acc (test) \\ \hline\hline
Emognition, valence & 38 / 67 / 72 / 106 / 36 & 11 / 24 / 17 / 22 / 11 & 0.282 \\
Emognition, arousal & 90 / 96 / 43 / 70 / 20 & 17 / 19 / 14 / 27 / 8 & 0.318 \\
MDMER, valence & 90 / 302 / 367 / 869 / 217 & 85 / 136 / 74 / 159 / 21 & 0.335 \\
MDMER, arousal & 300 / 652 / 378 / 430 / 85 & 63 / 142 / 134 / 125 / 11 & 0.299 \\
\Xhline{3\arrayrulewidth}
\end{tabular}}
\label{Tab:label_counts}
\end{table}

\paragraph{Repeated runs.}
\Tabref{Tab:multi_seed} reports the mean and standard deviation over three seeds for EVER and its EEG backbone.
EVER obtains a higher mean than CBraMod in all three metrics on both datasets.
An evaluation over multiple subject splits requires a full training run per split for every model, and the present benchmark is limited to a single subject split.

\begin{table}[h!]
\centering
\caption{\revb{\textbf{Mean $\pm$ standard deviation over three seeds (Acc / UAR / W-F1).}}}
\footnotesize
\resizebox{\linewidth}{!}{
\begin{tabular}{c|c|c}
\Xhline{3\arrayrulewidth}
Model & Emognition & MDMER \\ \hline\hline
CBraMod & 0.322 $\pm$ 0.030 / 0.283 $\pm$ 0.036 / 0.272 $\pm$ 0.033 & 0.295 $\pm$ 0.009 / 0.209 $\pm$ 0.003 / 0.234 $\pm$ 0.014 \\
\textbf{Ours} & \textbf{0.341 $\pm$ 0.026 / 0.310 $\pm$ 0.013 / 0.301 $\pm$ 0.008} & \textbf{0.321 $\pm$ 0.029 / 0.225 $\pm$ 0.014 / 0.269 $\pm$ 0.021} \\
\Xhline{3\arrayrulewidth}
\end{tabular}}
\label{Tab:multi_seed}
\end{table}

\paragraph{Paired comparison and logit-averaging ensemble.}
\Tabref{Tab:paired_bootstrap} compares EVER with its EEG backbone and with the logit-averaging ensemble of its two backbones on MDMER, using a two-sided paired bootstrap over the test trials.
\Tabref{Tab:ensemble} reports the ensemble for both backbone pairs on both datasets.
EVER is higher than the ensemble of its own backbones in every case, which indicates that the gain is not explained by averaging two unimodal predictions.

\begin{table}[h!]
\centering
\caption{\revb{\textbf{Paired comparison on the MDMER dataset.} The two-sided paired-bootstrap $p$-value of the difference from Ours is given in parentheses.}}
\footnotesize
\resizebox{0.9\linewidth}{!}{
\begin{tabular}{c|ccc}
\Xhline{3\arrayrulewidth}
Model & Acc (p-value) & UAR (p-value) & W-F1 (p-value) \\ \hline\hline
\textbf{Ours (VideoMAE+CBraMod)} & \textbf{0.354} & \textbf{0.237} & \textbf{0.280} \\
CBraMod & 0.302 (0.012) & \underline{0.208} (0.007) & \underline{0.219} (0.001) \\
Logit-averaging ensemble (VideoMAE+CBraMod) & \underline{0.306} (0.019) & \underline{0.208} (0.011) & 0.217 (0.001) \\
\Xhline{3\arrayrulewidth}
\end{tabular}}
\label{Tab:paired_bootstrap}
\end{table}

\begin{table}[h!]
\centering
\caption{\revb{\textbf{Logit-averaging ensemble versus EVER with the same backbones (Acc / UAR / W-F1).}}}
\footnotesize
\resizebox{0.85\linewidth}{!}{
\begin{tabular}{c|c|c|c}
\Xhline{3\arrayrulewidth}
Backbones & Model & Emognition & MDMER \\ \hline\hline
\multirow{2}{*}{CBraMod + AdaMAE} & Ensemble & 0.353 / 0.288 / 0.294 & 0.300 / 0.203 / 0.212 \\
& \textbf{Ours} & \textbf{0.371 / 0.306 / 0.298} & \textbf{0.341 / 0.231 / 0.274} \\ \hline
\multirow{2}{*}{CBraMod + VideoMAE} & Ensemble & 0.359 / 0.292 / 0.298 & 0.306 / 0.208 / 0.217 \\
& \textbf{Ours} & \textbf{0.371 / 0.353 / 0.353} & \textbf{0.354 / 0.237 / 0.280} \\
\Xhline{3\arrayrulewidth}
\end{tabular}}
\label{Tab:ensemble}
\end{table}

\paragraph{Fusion strategies with other backbones and datasets.}
\Tabref{Tab:fusion_more} extends the fusion comparison of \Tabref{Tab:ablation_fusion} to the VideoMAE backbone on Emognition and to both backbones on MDMER.
All fusion variants share the backbones, the learning rate, the schedule, and the class-weighted loss of EVER.
On Emognition, EVER is higher than concatenation, cross-attention, FiLM, and the pooled-node variant in all three metrics.
On MDMER, concatenation, cross-attention, and FiLM predict only the majority class under the shared hyperparameters (class 3 for valence and class 1 for arousal), as shown in \Figref{fig:cm_fusion}.
The three fusion operators therefore yield identical results that coincide with the majority-class predictor, whereas EVER exceeds this level in all three metrics for both backbone pairs.

\begin{table}[h!]
\centering
\caption{\revb{\textbf{Fusion strategies with identical backbones (Acc / UAR / W-F1).}}}
\footnotesize
\resizebox{\linewidth}{!}{
\begin{tabular}{c|c|c|c}
\Xhline{3\arrayrulewidth}
\multirow{2}{*}{Fusion} & Emognition & \multicolumn{2}{c}{MDMER} \\ \cline{2-4}
& VideoMAE + CBraMod & AdaMAE + CBraMod & VideoMAE + CBraMod \\ \hline\hline
Majority-class predictor & 0.300 / 0.200 / 0.139 & 0.317 / 0.200 / 0.153 & 0.317 / 0.200 / 0.153 \\
Concatenation & 0.300 / 0.265 / 0.280 & 0.317 / 0.200 / 0.153 & \underline{0.317} / 0.200 / 0.153 \\
Cross-attention & 0.324 / 0.323 / \underline{0.309} & 0.317 / 0.200 / 0.153 & \underline{0.317} / 0.200 / 0.153 \\
FiLM & 0.318 / \underline{0.330} / 0.298 & 0.317 / 0.200 / 0.153 & \underline{0.317} / 0.200 / 0.153 \\
Logit-averaging ensemble & \underline{0.359} / 0.292 / 0.298 & 0.300 / \underline{0.203} / \underline{0.212} & 0.306 / \underline{0.208} / \underline{0.217} \\
Pooled modality nodes & 0.324 / 0.277 / 0.288 & \underline{0.320} / 0.202 / 0.167 & 0.296 / 0.195 / 0.156 \\
\textbf{Ours (Inter-modal)} & \textbf{0.371 / 0.353 / 0.353} & \textbf{0.341 / 0.231 / 0.274} & \textbf{0.354 / 0.237 / 0.280} \\
\Xhline{3\arrayrulewidth}
\multicolumn{4}{r}{\textbf{Bold}: The best, \underline{Underline}: The second-best} \\
\end{tabular}}
\label{Tab:fusion_more}
\end{table}

\paragraph{Component ablation on Emognition and VCG prior on MDMER.}
\Tabref{Tab:ablation_components_emog} reports the component ablation of \Tabref{Tab:ablation_components} on Emognition with the AdaMAE and CBraMod backbones.
The BAR-GCN-only configuration and the full model obtain comparable accuracies of 37.6\% and 37.1\%, which correspond to a difference of one correct prediction over the two axes, and the full model obtains the highest UAR.
\Tabref{Tab:ablation_vcg_mdmer} repeats the ablation of the VCG prior in \Tabref{Tab:ablation_vcggcn} on MDMER, where removing either component lowers all three metrics.

\begin{table}[h!]
\centering
\begin{minipage}[t]{0.55\linewidth}
\centering
\caption{\revb{\textbf{Ablation study of EVER components on the Emognition dataset.}}}
\footnotesize
\resizebox{\linewidth}{!}{
\begin{tabular}{c|ccc|ccc}
\Xhline{3\arrayrulewidth}
\multirow{2}{*}{Modality} 
& \multirow{2}{*}{BAR-GCN} 
& \multirow{2}{*}{Clip nodes} 
& \multirow{2}{*}{VCG}
& \multicolumn{3}{c}{\textbf{Emognition}} \\ \cline{5-7}
& & & & Acc & UAR & W-F1 \\ \hline\hline
Video & & & & 0.312 & 0.255 & 0.231 \\ \hline
EEG & & & & 0.353 & 0.287 & 0.292 \\ \hline
\multirow{3}{*}{Video + EEG}
& \checkmark & & & \textbf{0.376} & \underline{0.295} & \textbf{0.314} \\
& \checkmark & \checkmark & & 0.353 & 0.293 & \underline{0.306} \\
& \checkmark & \checkmark & \checkmark & \underline{0.371} & \textbf{0.306} & 0.298 \\
\Xhline{3\arrayrulewidth}
\multicolumn{7}{r}{\textbf{Bold}: The best, \underline{Underline}: The second-best} \\
\end{tabular}}
\label{Tab:ablation_components_emog}
\end{minipage}
\hfill
\begin{minipage}[t]{0.43\linewidth}
\centering
\caption{\revb{\textbf{Ablation of the VCG temporal prior on the MDMER dataset.}}}
\footnotesize
\resizebox{\linewidth}{!}{
\begin{tabular}{c|c|ccc}
\Xhline{3\arrayrulewidth}
\multirow{2}{*}{Method} & \multirow{2}{*}{Variant}
& \multicolumn{3}{c}{\textbf{MDMER}} \\ \cline{3-5}
& & Acc & UAR & W-F1 \\ \hline\hline
\multirow{3}{*}{Ours (VCG prior)}
& Full & \textbf{0.341} & \textbf{0.231} & \textbf{0.274} \\
& w/o $\mathbf{p}_n$ & 0.308 & 0.197 & \underline{0.186} \\
& w/o $\phi_{nn'}$ & \underline{0.319} & \underline{0.202} & 0.165 \\
\Xhline{3\arrayrulewidth}
\multicolumn{5}{r}{\textbf{Bold}: The best, \underline{Underline}: The second-best} \\
\end{tabular}}
\label{Tab:ablation_vcg_mdmer}
\end{minipage}
\end{table}

\paragraph{Within-region attention weights.}
\Tabref{Tab:alpha_distribution} reports the within-region attention weights of \eqnref{eq:attention_pooling} on the test trials.
On MDMER, the weights are close to uniform in the frontal, parietal, and occipital regions and deviate moderately in the temporal and central regions, so the attention pooling does not single out individual electrodes.
On Emognition, where each region contains two electrodes, the weights vary across trials.
The contribution of BAR-GCN therefore originates from its anatomical structure rather than from electrode selection.
This interpretation is consistent with \Tabref{Tab:ablation_bargcn}, where replacing the anatomical partition with a random partition reduces the W-F1 from 27.4\% to 18.5\% and removing the intra-region mask or the inter-region edges lowers the accuracy to 31.2\% and 31.9\%.
The region nodes thus serve as anatomically grounded units that the inter-modal graph relates to the video clip nodes, which is the design objective of BAR-GCN.

\begin{table}[h!]
\centering
\caption{\revb{\textbf{Within-region attention weights on the test trials.}}}
\footnotesize
\resizebox{\linewidth}{!}{
\begin{tabular}{c|c|c|c}
\Xhline{3\arrayrulewidth}
Dataset (model) & Region & Electrodes & Mean weight per electrode \\ \hline\hline
\multirow{5}{*}{MDMER (VideoMAE+CBraMod)} & Frontal & 7 & 0.143 / 0.144 / 0.143 / 0.142 / 0.143 / 0.143 / 0.143 \\
& Temporal & 2 & 0.446 / 0.554 \\
& Central & 3 & 0.312 / 0.330 / 0.358 \\
& Parietal & 3 & 0.336 / 0.336 / 0.328 \\
& Occipital & 3 & 0.327 / 0.340 / 0.332 \\ \hline
\multirow{2}{*}{Emognition (AdaMAE+CBraMod)} & Frontal & 2 & 0.546 / 0.454 \\
& Temporal & 2 & 0.540 / 0.460 \\
\Xhline{3\arrayrulewidth}
\end{tabular}}
\label{Tab:alpha_distribution}
\end{table}

\paragraph{Confusion matrices.}
\Figref{fig:cm_main} shows the confusion matrices of CBraMod, VideoMAE, and EVER (VideoMAE+CBraMod) on the Emognition dataset.
On arousal, CBraMod assigns 73 of the 85 trials to a single class and VideoMAE never predicts the highest class.
EVER distributes its predictions over all five classes on both axes and obtains correct predictions in every class.
This suggests that the two modalities provide evidence for different classes and that both are required for balanced recognition.
\Figref{fig:cm_fusion} shows the confusion matrices of the fusion models on the MDMER dataset.
Concatenation, cross-attention, and FiLM assign every trial to the majority class, whereas EVER predicts multiple classes on both axes.
On this challenging task, a simple feature-level fusion of the two modalities collapses to the majority class, and the structured node-level interaction of EVER is required to obtain class-discriminative predictions.

\begin{figure}[h!]
\centering
\includegraphics[width=0.85\linewidth]{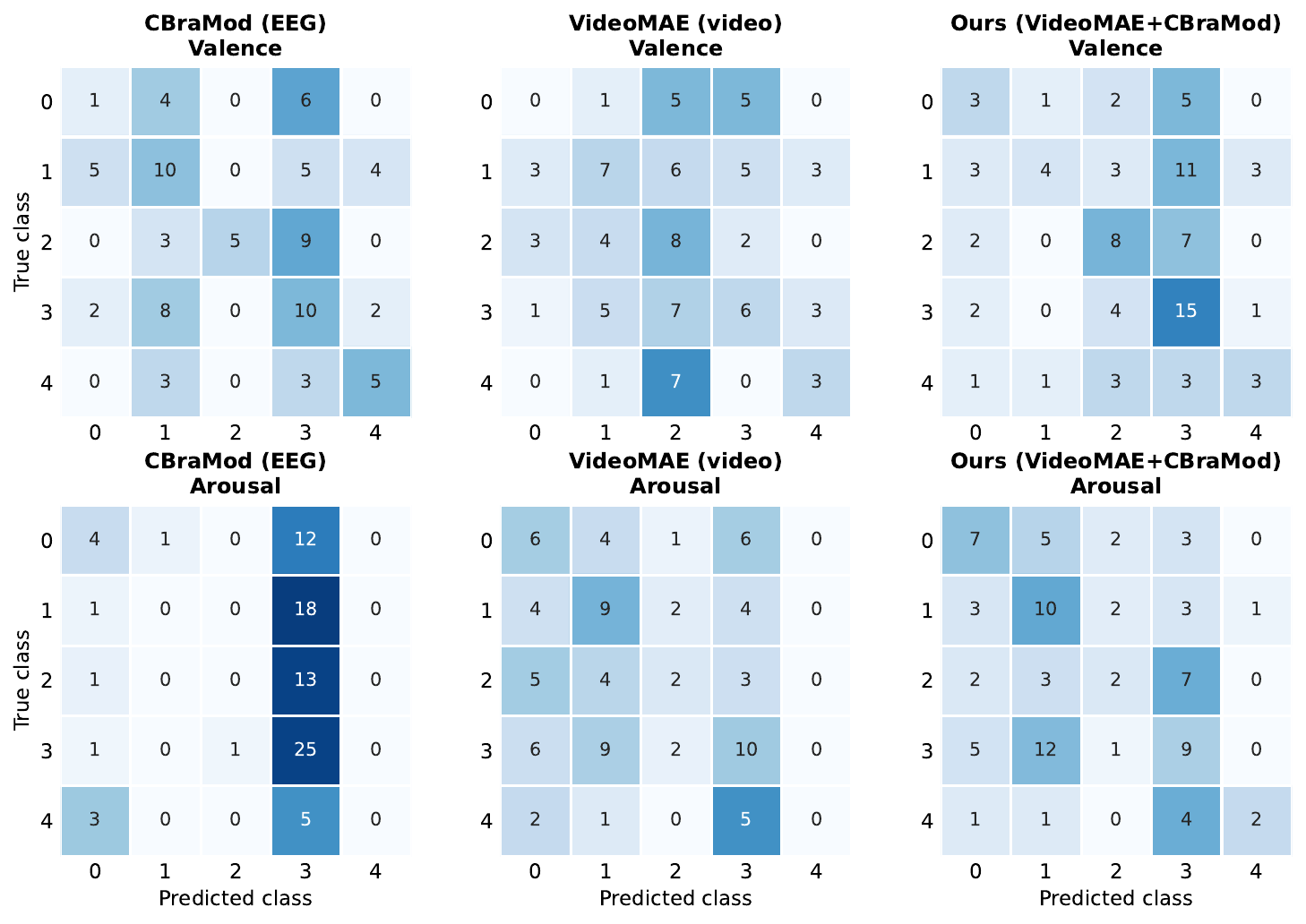}
\caption{\revb{\textbf{Confusion matrices of the main models on the Emognition dataset.} Each entry is the number of the 85 test trials.}}
\label{fig:cm_main}
\end{figure}

\begin{figure}[h!]
\centering
\includegraphics[width=\linewidth]{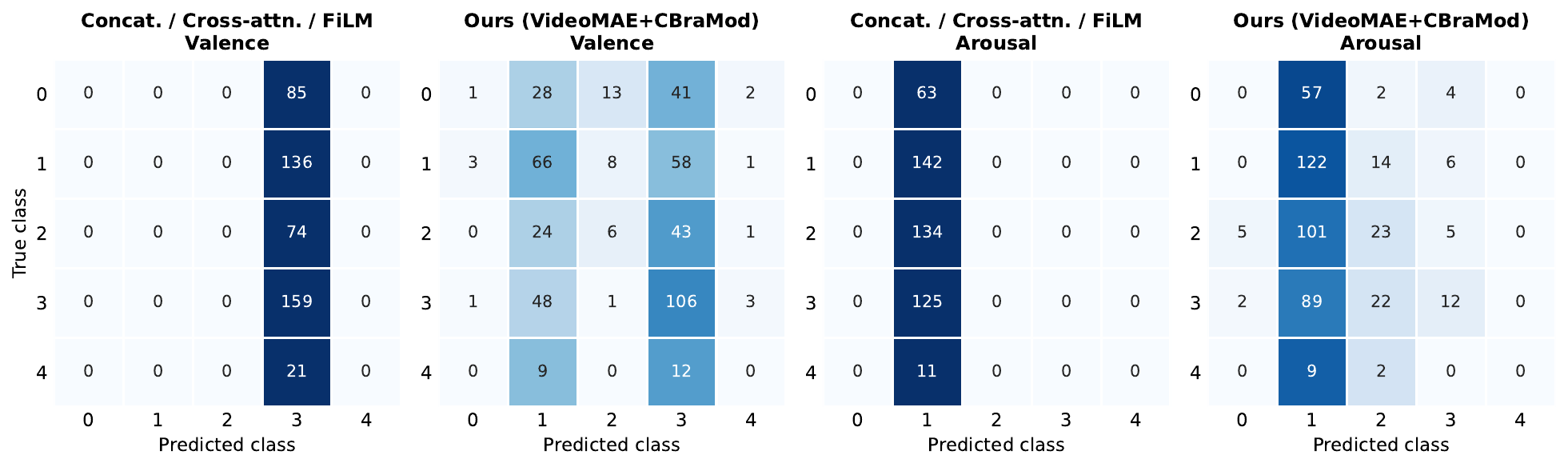}
\caption{\revb{\textbf{Confusion matrices of the fusion models on the MDMER dataset with VideoMAE + CBraMod.} Each entry is the number of the 475 test trials. Under the shared hyperparameters, the selected checkpoints of concatenation, cross-attention, and FiLM produce identical predictions and assign every trial to the majority class.}}
\label{fig:cm_fusion}
\end{figure}

}

\end{document}